\documentclass{WileyMSP-template}
\usepackage{acronym}
\usepackage{subcaption}
\usepackage{hyperref}
\usepackage{xurl}

\acrodef{SNN}[SNN]{Spiking Neural Network}
\acrodef{STDP}[STDP]{Spike-timing-dependent plasticity}
\acrodef{DG}[DG]{Dentate Gyrus}
\acrodef{PC}[PC]{Place Cell}
\acrodef{CA}[CA]{Cornu Ammonis}
\acrodef{LIF}[LIF]{Leaky Integrate-and-Fire}
\acrodef{EC}[EC]{Entorhinal Cortex}
\acrodef{ANN}[ANN]{Artificial Neural Network}
\acrodef{LTP}[LTP]{Long-Term Potentiation}
\acrodef{LTD}[LTD]{Long-Term Depression}
\acrodef{ANN}[ANN]{Artificial Neural Network}
\acrodef{PPC}[PPC]{Posterior Parietal Cortex}
\acrodef{SLAM}[SLAM]{Simultaneous Location And Mapping}

\begin{document}

\pagestyle{fancy}
\rhead{\includegraphics[width=2.5cm]{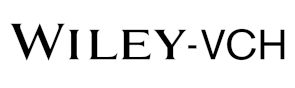}}

\title{Bio-inspired spike-based Hippocampus and Posterior Parietal Cortex models for robot navigation and environment pseudo-mapping}

\maketitle





\author{Daniel~Casanueva-Morato*\textsuperscript{,1}, Alvaro~Ayuso-Martinez\textsuperscript{1}, Juan~P.~Dominguez-Morales\textsuperscript{1,2}, \\Angel~Jimenez-Fernandez\textsuperscript{1,2}, Gabriel~Jimenez-Moreno\textsuperscript{1}, Fernando~Perez-Peña\textsuperscript{3}}

\begin{affiliations}
\textsuperscript{1} Robotics and Tech. of Computers Lab, Universidad de Sevilla, Spain\\
\textsuperscript{2} SCORE Lab, I3US, Universidad de Sevilla, Spain\\ 
\textsuperscript{3} School of Engineering, Universidad de Cádiz, Spain\\
Email Address: dcasanueva@us.es
\end{affiliations}


\keywords{spatial navigation, environment state map, Spiking Neural Networks, Hippocampus, Posterior Parietal Cortex, Neuromorphic engineering, SpiNNaker}

\begin{abstract}

The brain has a great capacity for computation and efficient resolution of complex problems, far surpassing modern computers. Neuromorphic engineering seeks to mimic the basic principles of the brain to develop systems capable of achieving such capabilities. In the neuromorphic field, navigation systems are of great interest due to their potential applicability to robotics, although these systems are still a challenge to be solved. This work proposes a spike-based robotic navigation and environment pseudomapping system formed by a bio-inspired hippocampal memory model connected to a Posterior Parietal Cortex model. The hippocampus is in charge of maintaining a representation of an environment state map, and the PPC is in charge of local decision-making. This system was implemented on the SpiNNaker hardware platform using Spiking Neural Networks. A set of real-time experiments was applied to demonstrate the correct functioning of the system in virtual and physical environments on a robotic platform. The system is able to navigate through the environment to reach a goal position starting from an initial position, avoiding obstacles and mapping the environment. To the best of the authors knowledge, this is the first implementation of an environment pseudo-mapping system with dynamic learning based on a bio-inspired hippocampal memory.


\end{abstract}
\section{Introduction}
\label{sec:introduction}

The brain, despite being one of the most complex and less known organs, has demonstrated great computation capabilities, adaptability and efficient resolution of complex problems, overcoming modern computers in many aspects \cite{budd2015early}. On this basis, neuromorphic engineering focuses on the study, design and implementation of hardware and software systems that mimic the principles of brain structures and their function to achieve the observed power efficiency and computational capabilities \cite{mead1990neuromorphic,indiveri2011neuromorphic}.

In order to develop systems that mimic these biological principles, a bio-inspired computational approach is needed. A specific type of biologically-plausible neural network, called \ac{SNN}, is the most widely used for this purpose. These networks, sets of neurons interconnected by synapses, are responsible for the generation, processing and transmission of asynchronous electrical pulses called action potentials or spikes. The transmission of spikes through different layers of neurons allow these systems to achieve high-level functionalities. Within the system, only the set of neurons that receive spikes are active, thus achieving a distributed computing approach that has great advantages in terms of energy consumption and real-time operation compared to traditional systems \cite{NeuromorphNature,zhu2020comprehensiveReview}.


In the neuromorphic engineering field, the navigation of animals and, particularly, mammals, through unknown environments is a topic of great interest due to its applicability to robotics. However, this is still an open challenge to be solved. In biology, the navigation process of mammals consists of different stages: mapping the environment, path planning and decision-making or motor control \cite{koul2019waypoint,tang2018gridbot}. Different brain regions are involved in the navigation process and each of its stages. Among them, the hippocampus and the \ac{PPC} should be highlighted. The hippocampus is the brain region that acts as a short-term memory capable of maintaining an allocentric representation of the environment map thanks to the place cells \cite{rolls2021brain,anand2012hippocampus}. On the other hand, the \ac{PPC} is the region responsible for, among other functions, the planning and modulation of the mammal's movements \cite{lyamzin2019mouse,andersen2002intentional,whitlock2008navigating}. 

Biological evidence suggests that the functioning and interconnection of these regions play an important role in the navigation process \cite{oess2017computational}. The hippocampus is involved in the exploration of the environment by maintaining an allocentric representation of the state of the environment, while the PPC is responsible for decision-making or translation of this allocentric information from the environment into a sequence of motor actions (egocentric) \cite{nakagawa2022neural, tosoni2014decision, oess2017computational}.

Regarding this topic, state-of-the-art works can be found in the literature. Some authors have attempted to develop navigation systems using spiking but not bio-inspired networks, such as \cite{fischl2017neuromorphic}. Others propose fully bio-inspired navigation systems (including all the stages in the process), such as \cite{oess2017computational}, which does not present any learning phase and, therefore, the environment is hard-coded in a static memory; and \cite{tang2018gridbot}, which is only able to map 3 different possible states.



Other works focus on a specific stage of the whole process, such as environment mapping \cite{nakagawa2022neural,kreiser2018pose,kreiser2020error,tang2019spiking}, path planning \cite{koziol2014neuromorphic,hwu2017adaptive,friedrich2016goal,koul2019waypoint,sakurai2021path}, and motor control \cite{zahra2021differential,nichols2010case}. In general, these works assume that the previous stages are global and return detailed and complete information that is complex to obtain in a spiking paradigm. Moreover, they operate globally rather than locally. Therefore, the mapping is done on the complete environment, the planning generates all the steps to reach the target position from the initial position and the motor control returns the exact sequence of movements to follow the complete route. These processes require very detailed information about the conditions of the complete environment, and they either are not always available or require information obtained through these same processes but locally.


Given these shortcomings and limitations, this paper proposes a fully-functional spike-based robotic navigation and environment pseudo-mapping system implemented with \acp{SNN} on the SpiNNNaker hardware platform. This system is capable of navigating while avoiding obstacles in an initially unknown grid environment to reach a goal position while mapping the environment with several possible states for each position. The operation of this system is achieved thanks to the use of a bio-inspired hippocampus memory model capable of generating and maintaining the environment map, and a bio-inspired \ac{PPC} model which is responsible for local decision-making or planning.

\subsection{Biological background}
\label{subsec:biological_background}

\subsubsection{Hippocampus}
\label{subsubsec:biologicalhippocampus}

The hippocampus or hippocampal system is that set of brain structures, belonging to the limbic system, which receives information from the different cortical sensory flows from the neocortex. All this information reaches the hippocampus mainly through the \ac{EC} (Brodmann's area 28) which, at the same time, acts as its main activity output pathway \cite{rolls2021brain}.

This region is characterized by acting as a short-term memory able to quickly store input information without following any structure. This learning and storage occurs with the association of the different input activity flows by means of an internal self-associative or attractor recurrent collateral network. As the input information is received from different cortical sensory streams, the association of this sensory information forms what is called a memory or episode. Thus, the hippocampus is involved in episodic memory \cite{rolls2021brain, anand2012hippocampus}.

Events stored in the episodic memory lack interpretation or reasoning, and are merely associations of information without cognitive processing. It is the semantic memory of the neocortex which gradually builds and adjusts, on the basis of much accumulated information, the semantic or cognitive representation of events \cite{rolls2021brain, anand2012hippocampus}.

Furthermore, the hippocampus consists of place cells, which are neurons that are activated when the subject is in a certain spatial region (not a specific point, but a specific area). A network of such neurons in the hippocampus provides a cognitive map of the entire environment in which the individual is located. This map stores an allocentric representation of both the environment and the location of the individual in that environment. It contains updated and coherent information regarding the state of the environment. This means that the hippocampus is involved in the navigation process through the environment \cite{rolls2021brain, andh12, whitlock2008navigating}.

The firing or activation field of the hippocampal place cells or, in other words, the state of the stored map, can be modified by both external signals (new sensory information from the environment) and internal signals (information regarding the individual's own movement) \cite{whitlock2008navigating}.

\subsubsection{Posterior Parietal Cortex}
\label{subsubsec:biologicalppc}

The \ac{PPC} (Brodmann areas 5, 7, 39 and 40) is an "associative" cortical region, since it is neither strictly sensory nor motor, but combines input information from several brain areas (such as somatosensory, auditory, visual, motor, cingulate and prefrontal) and integrates proprioceptive and vestibular signals from subcortical areas to achieve higher abstraction functions \cite{whitlock2017posterior}.

This anatomical situation within the brain makes it a very active region in several cognitive processes such as: sensorimotor integration, functional memory, imitation of actions observed in other individuals, early motor planning, transformation of sensory information into decisions and actions, and attention and spatial navigation, among others \cite{whitlock2017posterior, zhou2019posterior, lyamzin2019mouse}. Among all the cognitive processes, its contribution to navigation and the generation of actions based on sensory information should be highlighted.

During the subject's movement in an environment, \ac{PPC} neurons fire only when a particular movement or action (turning left, turning right, going straight ahead, etc.) is performed at particular (discrete) points along the pathway \cite{whitlock2017posterior, nitz2006tracking, andersen2002intentional, whitlock2008navigating}. In addition, these firing patterns are modulated from internal information and without the need for external visual stimuli \cite{whitlock2017posterior}.
These aspects support the fact that the \ac{PPC} plays a critical and essential role in the transformation of spatial information from landmarks based on allocentric vision (such as that present in the hippocampus) into sequences of movements based on egocentric (first-person) vision, called navigational context, which are necessary to reach the goal position  \cite{whitlock2017posterior, nitz2006tracking}.

Specifically, the activation of \ac{PPC} neurons encodes the movements to be performed before they take place, and is, therefore, responsible for the planning and modulation of the subject's movements \cite{lyamzin2019mouse, andersen2002intentional, whitlock2008navigating}. Regarding the internal landmarks, it is worth mentioning the use of the relative position of the subject's head within the allocentric map to determine the movement to be made to reach the target \cite{whitlock2017posterior, lyamzin2019mouse}.

The match-selective neurons present in the \ac{PPC} are responsible for planning the action or intention to be performed. For this purpose, they integrate information from sensory streams together with signals from brain regions involved in more abstract cognitive processes. These signals contain information related to internal brain models that identify the characteristics that are relevant from the point of view of the subject's behavior \cite{ibos2017sequential}. These neurons do not make any decision on the action to be taken until sufficient information is received \cite{lyamzin2019mouse}.

In short, there are neurons in the \ac{PPC} that receive the sensory information of the environment and the expected characteristics to carry out a certain behavior, thus, when the activation of both coincide, these neurons are activated, indicating the action to be performed.

These intentions are segregated in the \ac{PPC} in such a way that each region of the match-selective neurons specializes in planning different actions \cite{lyamzin2019mouse}. In addition, inhibitory neurons are activated in the \ac{PPC} between two consecutive decisions \cite{andersen2002intentional} to bring it to a starting state for the following decision.

\subsubsection{Hippocampus-Posterior Parietal Cortex relationship}
\label{subsubsec:relationppchipp}

The availability of a complete map of the environment thanks to the hippocampal system is not enough to perform spatial behaviors. A network capable of converting spatial information into goal-directed motor outputs is needed for this purpose. This is where the \ac{PPC}, which is in charge of transforming the allocentric representation of the environment stored in the hippocampus and \ac{EC} into goal-directed movements, comes into play. The connections between the hippocampus and \ac{EC} with the \ac{PPC} are responsible for supplying the \ac{PPC} with the spatial information needed for motor planning.

\subsection{Related work}
\label{subsec:related_work}

The navigation process in living beings presents different temporally separated phases of behaviour: environment mapping, path planning and motor control or translation of the action into motor commands \cite{koul2019waypoint, tang2018gridbot}.

Previous works present spiking navigation systems that cover these three phases, such as \cite{oess2017computational}. In that publication, the authors propose a navigation system based on 3 spatial frames of reference of the environment (allocentric, egocentric and route-centric) by means of an \ac{SNN} that consists of models of the hippocampus, the \ac{PPC} and the retrosplenial cortex. However, these models are static and do not have any learning capability. Therefore, they only work for completely predefined environments that are previously stored in the models. Furthermore, they are not fully bio-inspired models, since, although they mimic the high-level functionality of these brain regions, they do not rely on their neural structure and inner functioning to obtain these higher-abstraction functionalities.

In \cite{tang2018gridbot}, a complete robotic system based on \acp{SNN} capable of mapping unknown environments is proposed. Each map position is defined by the activation of a single place cell connected with dynamic synapses to a Border cell in order to detect the presence of obstacles, and also to a Goal cell to mark if it is the target. Therefore, the mapping of the environment is limited to 3 possible states per position: obstacle, target and none of the above. It also has other limitations, including the fact that movement planning is limited to avoiding obstacles and place cells are created on the fly for each new position that is visited. The latter behavior makes the system more optimized for representing the environment, although, at the same time, it moves away from the biological counterpart.

Other works address a particular phase of the navigation using \acp{SNN} and assume the rest as external inputs/outputs. 

For environment mapping, the publication by Nagagawa et al. \cite{nakagawa2022neural} proposes a system that stores the set of positions to visit in order to go from the initial position to the target, once it is located. This has also been addressed by \cite{kreiser2018pose,kreiser2020error,tang2019spiking}, who developed a spiking system capable of, approximately, determining the position of the robot and the presence of obstacles to achieve \ac{SLAM}.

Regarding the path planning phase, there are numerous works that focus on it. However, most of them assume that the map is static, that it has been completely defined previously and that the map is modelled using a 2D array of neurons, where each neuron represents a position and synapses represent the possibility of going from the source position to the target position.

Other works like \cite{koziol2014neuromorphic} and \cite{hwu2017adaptive} apply the wavefront technique both in simulation and in real environments to define the path between a source and a target position based on the set of projections that connect both with the smallest delay. A similar implementation can be seen in \cite{friedrich2016goal} (goal-driven), in \cite{koul2019waypoint} with neurons representing the points of interest instead of specific positions using a winner take all network, or in \cite{sakurai2021path} for dynamic maps with moving obstacles.

For the motor control phase, Zahara et al. \cite{zahra2021differential} propose a closed-loop control system for a robotic arm with an \ac{SNN} capable of translating desired actions into motor commands through a prior learning phase with \ac{STDP}, although it is not inspired in any brain region. In \cite{nichols2010case}, the authors design a complete navigation system; however, it focuses on the deployment of self-organizing \acp{SNN} to generate the appropriate motor command given an input of spatially distributed distance sensors. The system was not able to map the environment in memory and the path planning is based on following an object detected by the distance sensors. 

There are also other works, such as \cite{fischl2017neuromorphic}, that offer a different approach using \acp{SNN}, but they are quite distant from their biological counterpart and are rather focused on \ac{ANN} techniques. It makes use of a CNN network to generate the relevant motor commands via spiking-coded images from datasets.

\subsection{Main contributions}
\label{subsec:main_contributions}


The main contributions of this work include the following:
\begin{itemize}
    \item A spike-based bio-inspired \ac{PPC} neural model is proposed to determine and translate the motor action to be performed based on current information regarding the state of the local environment.
    
    \item A robotic system based on \acp{SNN} capable of navigating a grid environment to reach a target position while avoiding obstacles and mapping the state of the environment in real time is proposed. The system consists of a spike-based bio-inspired hippocampus memory model together with the proposed \ac{PPC} model.
    
    \item The proposed system is able to store a spiking representation of the map with the travelled path between a source and a target position together with information from the surrounding environment.
    
    \item The system was first simulated in software and then implemented on the SpiNNaker hardware platform and tested in real controlled environments with a mobile robotic platform.
    
    
    \item The source code is publicly available (see Section~\ref{sec:conclusions}), together with the documentation including all the necessary details regarding the \ac{SNN} architectures. 
\end{itemize}

The rest of the paper is structured as follows: Section~\ref{sec:materials_and_methods} presents the materials and methods used in this work. In Section~\ref{sec:computational_model}, the proposed spike-based robotic navigation and environment pseudo-mapping system is detailed, including its architecture (Section~\ref{subsec:architecture}) and its operating principle (Section~\ref{subsec:working_principle}). The experiments performed to evaluate the functionality and performance of the proposed PPC model and the complete system are explained in Section~\ref{sec:experiments_and_results}, along with the results obtained. Then, in Section~\ref{sec:discussion}, the results of the experiments are discussed. Finally, the conclusions of the paper are presented in Section~\ref{sec:conclusions}. 
\section{Materials and methods}
\label{sec:materials_and_methods}

\subsection{Spiking Neural Networks}
\label{subsec:spiking_neural_networks}

The third generation of neural networks, \acp{SNN}, is one of the best alternatives to use when working with bio-inspired computational models. These allow creating networks of neurons that are inspired in biology and aim to mimic it in order to incorporate the neurocomputational capabilities found in nature \cite{ahmed2020brain}.

The operation and the transmission of information of \acp{SNN} are based on spikes, which are asynchronous events that incorporate the concept of space and time into neural networks through connectivity and plasticity. These networks are very efficient from a computational point of view, as only the components that are involved in a specific event that occur at a specific moment of time (when input spikes are received) operate, due to their asynchronous behavior \cite{lobo2020spiking}.

At the biological level, there is a wide variety of models for each of the basic components that are present in a neural network depending on the level of abstraction and, therefore, the level of neurocomputational complexity to be achieved. Among them, the most widely-used one in the state of the art, the \ac{LIF} neuron model, was used in this work, which describes the behavior of a neuron by means of an RC circuit with a spike generator. \cite{tavanaei2019deep, stein1965theoretical}. On the other hand, \ac{STDP} learning mechanisms were also used, in which the weight of synapses are modified in proportion to the degree of temporal correlation between the activity of pre- and post-synaptic neurons \cite{sjostrom2010spike}.

Different hardware platforms particularly designed for implementing and simulating \acp{SNN} can be found in the literature. Some of the most well-known hardware platforms are SpiNNaker \cite{furber2014spinnaker}, Loihi \cite{davies2018loihi} and TrueNorth \cite{merolla2014million}. 

In this work, we used SpiNNaker as the hardware platform in which the different \ac{SNN} models presented were implemented and emulated. This platform works with a time step of 1~ms time unit and allows the modeling of large \acp{SNN} in real time. In addition, a software package called sPyNNaker \cite{rhodes2018spynnaker} allows running PyNN \cite{davison2009pynn} simulations directly on the SpiNNaker board, making the platform very straight-forward to work with, since all the codes regarding the design and implementation of \acp{SNN} can be done using high-level functions described in Python programming language.

\subsection{Robotic platform}
\label{subsec:robotic}

\begin{figure}[ht]
    \centering
    \includegraphics[width=0.4\textwidth]{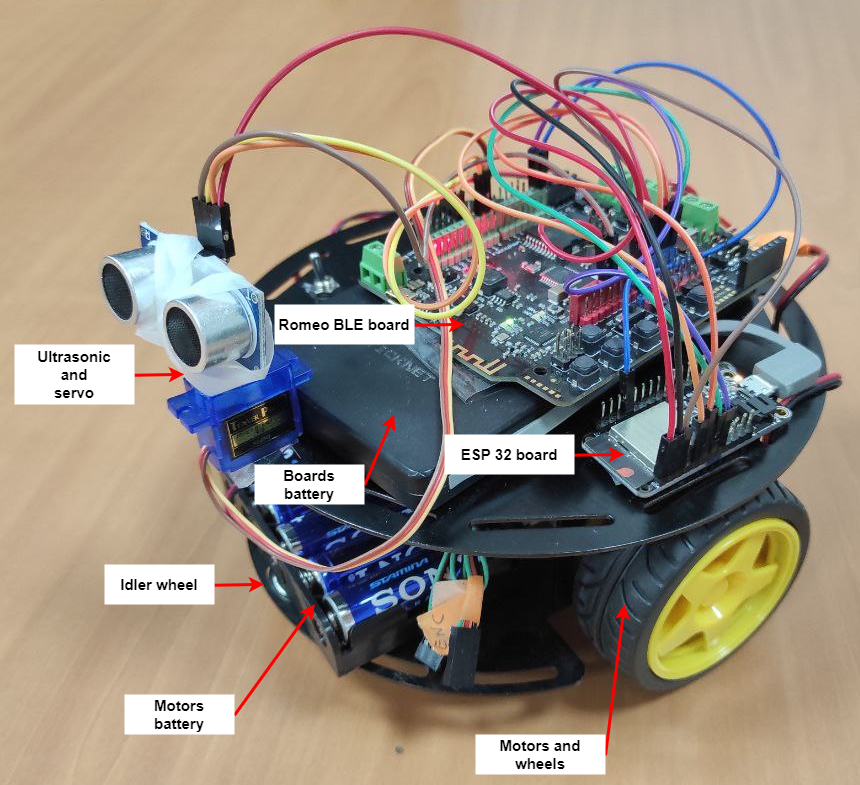}
    \caption{Picture of the robotic platform used for the real-time demonstration. The relevant components are highlighted: the chassis, the power supply system, the robot's control and communication boards, and the sensors and actuators needed to obtain local information from the environment.}
    \label{fig:robot}
\end{figure}

To test the correct operation of the proposed system in a real, physical environment, a customized robotic platform was used (Figure~\ref{fig:robot}). This robotic platform is capable of moving through the environment, obtaining information regarding the distance to nearby obstacles in order to detect them, and communicating with other systems wirelessly.

To achieve these functionalities, the platform consists of a circular chassis with 2 motorized wheels, an idler wheel and an ultrasound sensor mounted on a servomotor to be able to measure the distance to nearby obstacles in different directions around the robot without the need to physically turn. At the electronic level, it contains a Romeo BLE board and an ESP32 board. 

The Romeo BLE board is in charge of managing the robot control system. On the one hand, it takes distance measurements on the right, front and left of the robot through the ultrasound sensor with the help of the servomotor to detect whether or not there are obstacles in the near environment. This information from the local environment is then sent to the spiking system emulated in SpiNNaker for mapping the environment in memory. On the other hand, it has an integrated H-bridge motor controller to perform the motor actions received from the spiking system.

The ESP32 board is connected to the Romeo BLE board and has a wireless communication module. It is responsible for the wireless communication between the robotic platform and the spiking system. It creates an UDP server to which the spiking system must connect in order to transmit information. This allows a fast sending data from the Romeo BLE board to the spiking system and vice versa.

As a wireless robotic platform, it requires a portable power supply. On the one hand, the motors were powered using a set of 4 AAA batteries, while the rest of the electronic systems were powered with an external battery.
\section{Environment pseudo-mapping and navigation model based on Hippocampus and Posterior Parietal Cortex}
\label{sec:computational_model}

This paper proposes a spike-based robotic navigation and environment pseudo-mapping system consisting of a bio-inspired hippocampal memory model and a bio-inspired \ac{PPC} decision-making model that are interconnected with each other (see Figure~\ref{fig:arch}). This spiking navigation system follows the biological principles observed in living beings, where the joint functioning of both brain regions plays a critical and essential role in this process as discussed in Section~\ref{subsubsec:relationppchipp}.

The purpose of the system is to be able to reach a goal position within a grid environment while avoiding obstacles and walls that may be in the trajectory and, at the same time, map the state of the explored fragment of the environment in real time. The hippocampus is in charge of mapping the state of the environment by maintaining a spiking and updated representation of it during exploration, while the \ac{PPC} is responsible for the decision-making regarding the next action to perform based on the information obtained from the environment that is provided by the hippocampus.

The architecture of the models and, thus, of the navigation system, is fully parameterized and depends on the environment to be navigated and the number of possible states that the environment may have. Based on these two variables, the sizes of the different populations as well as the set of internal and external connections and interconnections are automatically calculated.

At the beginning, the environment is completely unknown. The system will map the environment as it is explored and, after reaching the target, maintain a stored representation of the state of the regions it has explored to reach that position. However, it is firstly necessary to specify the size of the grid in which the environment will be divided. This is necessary to define the possible positions within the environment and, thus, the size of the state map that the hippocampus model will store. The proposed architecture does not limit the maximum grid size.

\subsection{Architecture}
\label{subsec:architecture}

\begin{figure}[!t]
    \centering
    \includegraphics[width=0.9\textwidth]{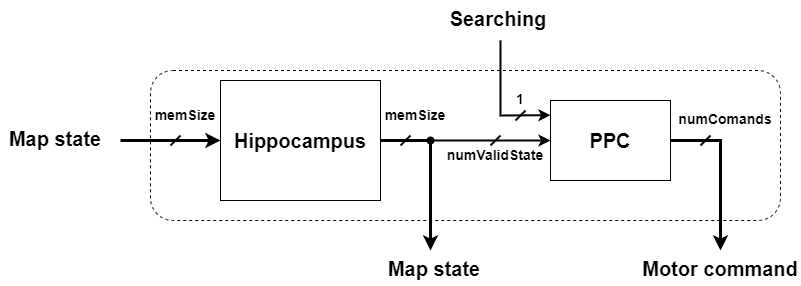}
    \caption{Architecture of the pseudo-mapping and navigation model proposed. The blocks that correspond to the architecture are framed with a dashed line, which consists of the hippocampus model and the \ac{PPC}. The architecture is parameterised in size through memSize (size of the memory, equivalent to the number of states plus the number of neurons needed to encode in binary all the positions of the map), numValidState (number of states indicating that the remembered position is traversable) and numCommands (number of motor commands that the PPC can generate). All the interconnections between these blocks are excitatory, static and parameterized according to the environment in which it will be used.}
    \label{fig:arch}
\end{figure}

The complete bio-inspired spike-based navigation system is shown in Figure~\ref{fig:arch}. Both the hippocampus model and the \ac{PPC} model are detailed in Section~\ref{subsubsec:hippocampus} and~\ref{subsubsec:ppc}, respectively.

\subsubsection{Hippocampus}
\label{subsubsec:hippocampus}

The bio-inspired spike-based hippocampus memory model is based on that reported in our previous work \cite{casanueva2022bio}. This memory model is able to learn and recall memories following a workflow and structure based on the biological model of the hippocampus. A memory is considered to be a pattern of spiking activity that is fed to the memory model that consists in the concatenation of information from different sources. 

The information contained in memories lacks prior interpretation, depends on the source it is provided from, and presents a spatio-temporal coding. This means that the activity of different neurons at the same instant belongs to the same memory, and that the activity of the same neuron at different instants refers to different memories.

In order for the learning and recalling operations to be performed, memories must contain a cue. This way, the learning operation stores the memory by means of the association of the cue to the rest of the content of the memory, while the recalling operation only requires the input cue to return the rest of the content of the memory with which it was associated. The memory also features a forgetting mechanism that, when a new memory is learned whose cue is the same as that used by another memory, it is able to forget the previous memory and store the new one.

Moreover, as in the biological model, since the hippocampus acts as a short-term memory, its content is forgotten over time (memory leak). By using the \ac{STDP} learning mechanism, for each operation in which a memory is not involved, the weight of the synapses in which it is stored are slightly decreased. Therefore, if after a configurable number of memory operations the memory is not used, the hippocampus model ends up forgetting it, as it is considered not important. Although the forgetting of the memory depends on the number of consecutive operations in which it is not involved, this fact is directly related to the time it has been stored without being used, hence the reason for calling it a temporal forgetting mechanism.

The memories that the hippocampus model learns, recalls and forgets take a position in the environment as a cue and the state of the environment at that position as the rest of the content. Taking these considerations, the memories would represent maps of the state of the environment in which the subject is located. Following the biological basis, the hippocampus model acts as a short-term memory that, thanks to the use of place cells, is able to maintain an allocentric representation of the map of the environment in which the subject is located \cite{oess2017computational}. Place cells would be responsible for storing memory cues.


\subsubsection{Posterior Parietal Cortex}
\label{subsubsec:ppc}

\begin{figure}[htbp]
    \centerline{\includegraphics[width=0.4\textwidth]{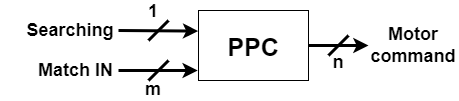}}
    \caption{Block diagram of the \ac{PPC} model including parameterized inputs and outputs. The \textit{Searching} input signal indicates the start of the motor command planning process whose signal will be propagated internally over time. \textit{Match IN} are the input signals in which the activation of any of them in conjunction with the \textit{Searching} signal determines the command to perform, i.e., the Motor command output.}
    \label{fig:ppcbasic}
\end{figure}

\begin{figure}[!t]
    \centering
    \includegraphics[width=0.9\textwidth]{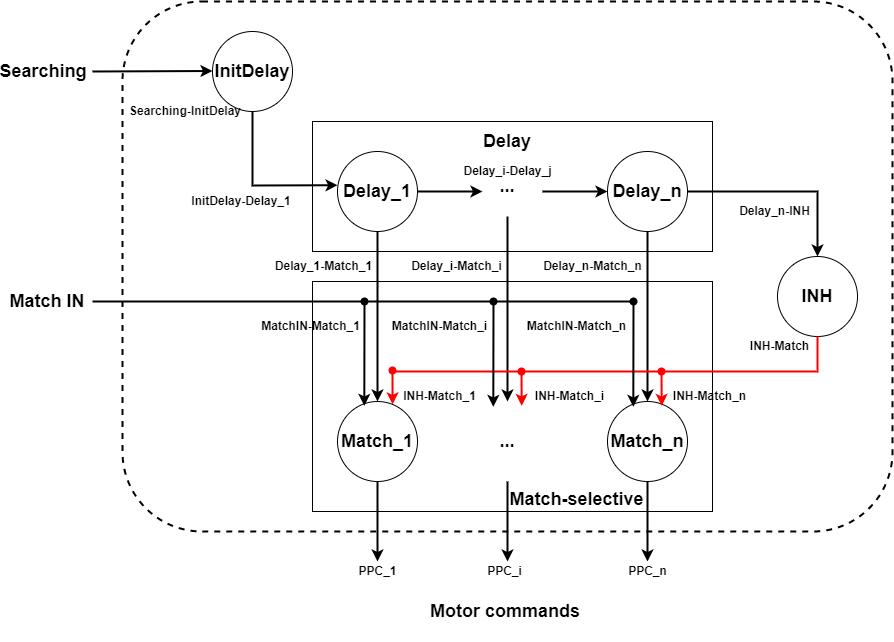}
    \caption{Internal structure of the \ac{PPC} model. It consists of 4 populations of neurons: \textit{InitDelay}, \textit{Delay}, \textit{INH} and \textit{Match-selective}. The \textit{InitDelay} and \textit{Delay} populations act as delayed propagators of the input signal, the \textit{Match} population determines whether the \textit{Delay} activation matches the inputs, indicating the action to be performed, and the \textit{INH} population is responsible for regulating the output activity of the network. Synapses denoted with black arrows are excitatory and those presented with red arrows are inhibitory.}
    \label{fig:ppcdetail}
\end{figure}

The proposed computational model of the \ac{PPC} is based on the biological basis presented in Section~\ref{subsubsec:biologicalppc}. This model is responsible for decision-making; this is, local planning and the generation of the robotic system's movements. By local planning, it is meant that the system decides, based on the information of the local state of the environment, what is the next action to be performed in order to reach the target position.

As shown in Figure~\ref{fig:ppcbasic}, to perform this task, the PPC takes time-coded local environmental state information from the hippocampus (Match IN) and a signal that identifies the start of the next decision-making (Searching) operation as input. From this allocentric information of the environment, the model returns the next motor command (egocentric information) to be performed in order to achieve the goal. 

As shown in Figure~\ref{fig:ppcdetail}, the signal that marks the beginning of the decision-making process (Searching) connects to the population of Delay neurons. This population consists of a set of neurons connected in a chain that is responsible for propagating the input signal with a certain delay. There is a time difference between the start of the decision-making process and the input of the first signal corresponding to the state of the first position of the local environment. As a consequence, the Delay population has an InitDelay input neuron with a different time delay than the rest, which aims to correct this time difference. 

Each Delay neuron is connected to a single Match-selective neuron. Thus, the population of Delay neurons would act as a selector for a single Match-selective neuron for each time window. The time that elapses between the selection of one match-selective neuron and the next depends on the propagation delay of synapses in the Delay neuron chain.

When the decision-making process starts, the possible states of interest of the different positions of the local environment of the robotic system arrive to the population of Match-selective neurons encoded in time. In other words, the network receives a set of inputs that identify the different states that are taken into account in a position to plan the next motor command. With time coding we mean that the activation of these signals at different time instants identify different positions within the local environment. 

The time delay in signal transmission used in the Delay population is the same as the time taken between the temporal state encoding of one position and the next. When the activation of both temporal inputs coincide, a specific Match-selective neuron is activated. As each Match-selective neuron encodes a single motor command, it is these neurons which, when activated due to the sensory information from the local environment, indicate the action to be performed.

In addition, the model presents a population of interneurons, \textit{INH}, which inhibit the neurons of the Match-selective population and receive an excitatory input from the last neuron of the Delay population. Therefore, the function of these neurons is to regulate the output activity of the model, leaving the Match-selective neurons in a resting state between one decision-making and the next, as observed in the biological basis. The rest of the synapses in the model are excitatory.

\subsection{Operating principle}
\label{subsec:working_principle}

The proposed spike-based robotic system is bio-inspired in the navigation system formed by the hippocampus and the \ac{PPC}. This system enables the navigation of the robot through a grid environment, initially unknown, to reach a goal position avoiding possible obstacles that may appear in the trajectory. At the same time, the system performs a pseudomapping of the state of the environment, allowing to know the current local environment traveled so far in order to determine the actions or motor commands to be performed to reach the goal. 

The term pseudomapping is used for two main reasons. The first reason is that the system does not traverse the environment to map it completely: it obtains a map with the state of the fragment of the environment that has been traversed to reach the target position. This map is formed by the combination of the local mapping performed at each position of the trajectory in order to plan the next motor action. The second reason is that the environment is not mapped based solely on the presence of obstacles and targets, as the mapping also considers the input of more complex states obtained by brain systems with higher abstraction functions.

For this purpose, the operation of the system is divided into iterations. In each iteration, the system maps the state of the local environment and, based on this information, determines the next action to be taken. Each of these processes is explained in detail below.

\subsubsection{Local environment mapping}
\label{subsubsec:mapping}

\threesubsection{\textbf{State map}}\\
The bio-inspired hippocampal memory model would be responsible for maintaining the map representation with the state of the environment traversed so far. Each position of the map is represented by the activation of a single \ac{PC} and, in this model, only one \ac{PC} is activated simultaneously thanks to the internal mechanism of input information dispersion \cite{casanueva2022bio}. The activation of the \acp{PC} will be used as a cue to the memory to be learned and stored, while the remaining content of the memory corresponds to the possible states that these positions may have. 

The \ac{PC} map stores a total of 8 possible states per cell. Each \ac{PC} (i.e., each position on the map), will be associated with the activation of a single state. The main states are: obstacle, goal, free (non obstacle), start position and unexplored. The remaining three (step in path, crossroad and dead end) are derived from these five and are discussed in more detail below.

The unexplored status is used to identify those regions that have not been explored yet. The states referring to the presence or absence of obstacles are given by the activity input from Boundary Cells present in the Entorhinal Cortex. These neurons are activated when boundaries are detected in the environment either in the form of vertical surfaces or falling edges, i.e., in the presence of obstacles \cite{poulter2018neurobiology}. On the other hand, the goal state is given by the activation of Goal cells present in the Prefrontal Cortex. These neurons are activated when facing a position that corresponds to the goal \cite{tang2018gridbot}.

The activity of Boundary Cells and Goal cells also reaches other brain regions with higher levels of abstraction functions. In these regions, information is converted into knowledge, as occurs in the semantic memory \cite{rolls2021brain}. These areas would be responsible for the activity related to the remaining states that are stored in the positions of the environment map: step in path, crossroad and dead end. 

The \textit{step in path} state indicates that that specific position is a position that has either been passed through or, after the decision-making phase, established as the next position in the path. The \textit{crossroad} state identifies a position that has been passed through and where there is at least one additional obstacle-free neighboring position in addition to the one taken in the current path. This state is useful when backtracking during navigation after having encountered a dead end in the path to the goal. Finally, the \textit{dead end} state indicates which positions that have been traversed lead to dead ends in the trajectory to reach the goal position.

The main states (obstacle, goal, free, start position and unexplored) are determined internally in the network, while the states derived from them (step in path, crossroad and dead end) require higher processing. Since the aim of this system is not to compute the most complex states present in biological systems, they are processed externally to the network and fed into the memory afterwards. 

\threesubsection{\textbf{Implemented functionality}}\\
In the first phase, the system obtains from the real environment the state of the 3 neighboring positions (right, front and left) with respect to the current position of the robot and stores them in the hippocampal memory. To obtain the neighboring positions, the system combines the orientation and the current position of the robot. The orientation would come from the Head direction cells \cite{poulter2018neurobiology} and the position would come from the hippocampal place cells \cite{tang2018gridbot}.

The possible states of the real environment that reach the hippocampus would be: position with obstacle, position without obstacle and target, as the remaining states would come from  further processing. The only exception is the initial position state, which is manually specified in the system at the beginning of the simulation to position the robot within the environment. This information also reaches the \ac{PPC} where the decision-making phase takes place.

All learning operations applied on the hippocampal memory during the local mapping phase are performed with reinforcement. The hippocampal memory has a forgetting mechanism, which leads to forgetting information that has not been accessed after many consecutive operations. Therefore, as the environment is being navigated, the system will work with nearby locations, while distant ones will be in danger of being forgotten, as they have not been used for several operations. Reinforcement learning consists in a learning operation followed by a recalling operation of the same memory in order to store it for a much longer time.

\subsubsection{Decision-making}
\label{subsubsec:decision_making}

In this phase, the \ac{PPC} performs local decision making and planning, i.e., determining which motor action to perform in order to reach the goal based on the surrounding environment. In this way, the output of the hippocampal memory is connected to the input of the \ac{PPC}. Specifically, the signals from hippocampal states of interest that are passed to the match-selective neurons of the \ac{PPC} are: \textit{goal}, \textit{step in path} and \textit{crossroad}. This means that the system will determine the motor actions needed in order to reach the positions that present only those states. 

By default, the behavior of the system is to search for the next position to reach the target. To do so, it obtains the state information of the positions local to the current one from the hippocampus. If any of the neighboring positions present any of the states of interest, the \ac{PPC} generates the motor action needed to reach it. At the same time, if the current position has more than one neighboring position without obstacle, its state would be updated to \textit{crossroad}. 

If no motor action is generated (no goal, step in path or crossroad states found in its local environment), a global planning that is external to the network would be called to determine, from all the non-obstacle positions around the current one, the closest to the goal based on the Manhattan distance, and its state would be modified in memory to \textit{step in path}. In the next iteration of the system, the \ac{PPC} should find at least one position with a state of interest.

When checking the positions, if it is completely surrounded by obstacles except for the previous position through which it arrived at the current one, it means that it has encountered a dead end. In this case, the system would switch to a backtracking behavior. In this behavior, the system would turn around and start backtracking until it encounters a position with \textit{crossroad} state. Since it is a \textit{crossroad}, this position will have at least one other possible neighboring position with no obstacle, thus, the system will again continue to search for a path to the target starting from there. While performing the backtracking, it changes the \textit{step in path} positions to \textit{dead end} and the \textit{crossroad} positions that do not have any unexplored neighboring positions with no obstacles to \textit{step in path}. 

The system will finish its operation when reaching the target position.


\section{Experimentation and results}
\label{sec:experiments_and_results}

A set of experiments were performed to demonstrate the correct functioning of the system. 


For the case of the memory model, several experiments were performed in a previous work that validated its operation \cite{casanueva2022bio}. In this paper, it was implemented with the sPyMem Python library\footnote{\url{https://github.com/dancasmor/sPyMem}}. 

After that, an incremental experimentation focused on the whole system was performed. These experiments consisted in the navigation and mapping of a set of virtual grids of different sizes and with different distribution of obstacles. In these virtual environments, the robotic system was not used, and, therefore, data inputs from the environment were simulated and given as input to the system. 

Initially, the robotic system was evaluated on the navigation and mapping of small environments with few obstacles as a proof of concept. Subsequently, after demonstrating its operation in simple environments, larger environments with more strategically placed obstacles were used in order to test the system in different situations. After deeply evaluating the system in a virtual environment, a demonstration was performed on a physical environment using the robotic system described in Section~\ref{sec:materials_and_methods}.

All experimentation on the complete system in both virtual and real environments took place in real time and on a 2-dimensional plane. Due to these characteristics of the environment, a total of 4 motor actions are sufficient to move the robot through it, or, in other words, a total of 4 Match-selective neurons in the \ac{PPC} model are needed to cover the spectrum of possible motor actions. These actions would be: moving up, down, left and right.

To determine the motor action to be performed, a recall of the states of the neighboring positions is configured in such a way that the order of the input positions' state to the \ac{PPC} is: up, left, down and right. In addition, the \ac{PPC} model is defined with a delay in the propagation of the onset signal of the decision-making phase equal to the time it takes to perform a recall in the memory model. By taking both considerations, the activation of the 4 Match-selective neurons will indicate the motor actions required for moving up, left, down and right, respectively.

In addition, for these experiments, a numerical code is used to represent the states of the positions. This number represents the id of the hippocampal memory neuron whose activity encodes that particular state.

\subsection{PPC proof of concept}
\label{subsec:ppcpoc}

\begin{figure}[!t]
    \centering
    \includegraphics[width=0.99\textwidth]{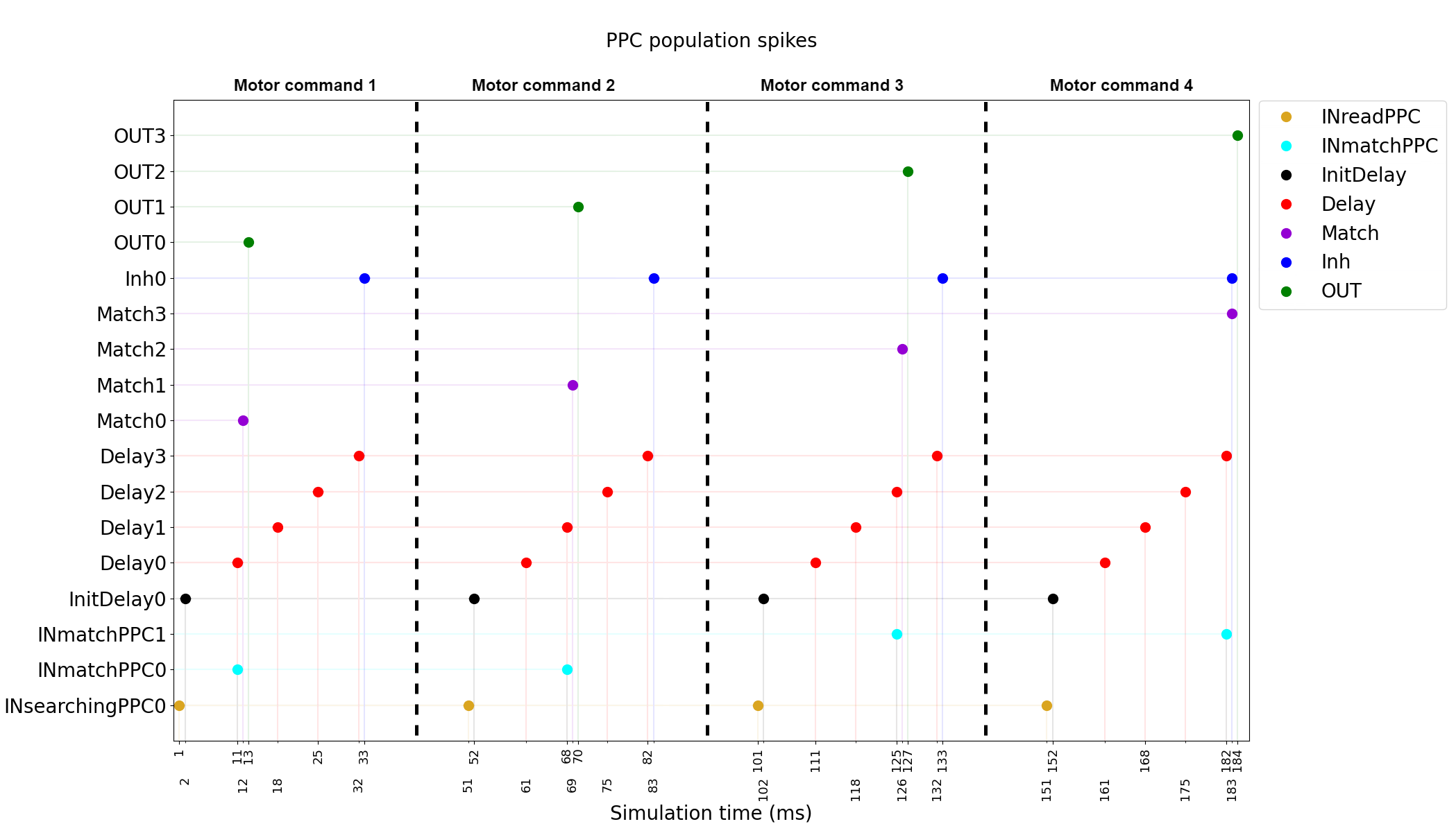}
    \caption{Internal and external spiking activity of the proposed \ac{PPC} model during a proof of concept test for the 4 possible motor commands. Each point represents the activation of a neuron (Y axis) at a given time instant (X axis). The plot is divided into four parts, each demonstrating the selection of a different motor command. Each part starts with the activation of the decision-making signal (\textit{INsearchingPPC0}), which is propagated along each neuron of the Delay population with a certain delay. When the activation of any of the input signals of interest (\textit{INmatchPPCi}) coincides with the activation of a Delay neuron (\textit{Delayj}), the corresponding Match neuron (\textit{Matchj}) is activated and, thus, a motor command (\textit{OUTj}) is selected.}
    \label{fig:ppctest}
\end{figure}

This experiment is a proof of concept to demonstrate the correct functioning of the PPC model. For this purpose, a PPC model capable of reacting to 2 possible input states (\textit{INmatchPPCi}) and, from them, generating 4 possible motor actions (\textit{Matchi}) was developed. This model has a signal propagation delay of 7 milliseconds within the Delay population and 9 milliseconds as the initial delay (\textit{InitDelay}). In other words, it takes 9~ms from the beginning of the decision-making phase to the arrival of the state of the first position, and a time of 7~ms between state arrivals of the remaining consecutive positions.

The test consists in carrying out 4 decision-makings, in each of which a different action is determined by the arrival of one of the two states of interest. Specifically, the model should select the first (upward), second (left), third (downward) and fourth (right) actions for the first, second, third and fourth decisions, respectively. Following the analogy discussed at the beginning of the section, the inputs from the states of interest will arrive temporarily in the following order: up, left, down and right.

Figure~\ref{fig:ppctest} shows the input and output spiking activity of the model, as well as the spiking activity of each neuron of each internal population of the network that is activated as a consequence for every time step of the simulation. 

The first decision-making (ms~1), starts with the arrival of a spike from the \textit{INsearchingPPC0} input. This spike will trigger the activation of the Delay population input neuron, \textit{InitDelay}, at ms~2. This neuron starts the propagation of the input signal along the Delay neuron chain. The first activation, \textit{Delay0}, occurs at ms~11 after 9~ms of delay, while the remaining neurons in the chain (\textit{Delay1}, \textit{Delay2} and \textit{Delay3}) will be activated every 7 ms (ms~18, 25 and~32, respectively). This temporal deviation between the activation delay of the first neuron and the remaining neurons is necessary to correct the temporal differences that exist between the beginning of the decision-making process and the arrival of the state of the first position of the local environment.

At the same time, at an instant in the decision-making process, one of the two inputs related to the states of interests for the position being evaluated will be activated. The activation of any of the two is sufficient to select the action to be taken. Specifically, for the first decision-making, the input of the first state of interest, \textit{INmatchPPC0}, is activated at ms~11. Since it coincides with the activation of the first neuron of the Delay chain, \textit{Delay0}. The first Match-selective neuron, \textit{Match0} is then activated at ms~12. In other words, the action of moving up is selected, whose decision will reach the output population of the model at ms~13.

After the activation of the last neuron of the chain in the Delay population (\textit{Delay3}) at ms~32, the decision-making phase is finished. Therefore, this activity is then propagated to the inhibitory interneuron \textit{Inh0}, whose activation at ms~33 will inhibit the Match-selective neurons in order to prepare them for the next decision-making.

What was described for the first decision-making, occurs for the following ones at ms~51, 101 and 151 respectively. The signal of the first state of interest (\textit{INmatchPPC0}) is activated for the second decision-making at ms~68, and the signal of the second state of interest (\textit{INmatchPPC1}) for the third and fourth decision-makings at ms~125 and 182, respectively. This activity coincides in time with those of the \textit{Delay1}, \textit{Delay2} and \textit{Delay3} neurons, respectively, leading to the activation of the \textit{Match1} (ms~69), \textit{Match2} (ms~126) and \textit{Match3} (ms~183) neurons. This activation of Match neurons results in the selection of the second action for the second decision-making, the third action for the third decision-making and the fourth action for the fourth decision-making, as reflected in the population external to the PPC model (OUTi) at ms 70, 127 and 184, respectively.

With this experiment, the correct functioning of the model in the selection of the motor actions for the different possible input conditions in the decision-making process was verified. For each decision-making process, the model was able to generate the expected motor commands after providing the given data inputs.

\subsection{4$\times$4 grid map simulation experiment}
\label{subsec:4x4}

\begin{figure}[!t]
    \centering
    \begin{subfigure}{.33\textwidth}
    \centering
    \includegraphics[width=0.99\textwidth]{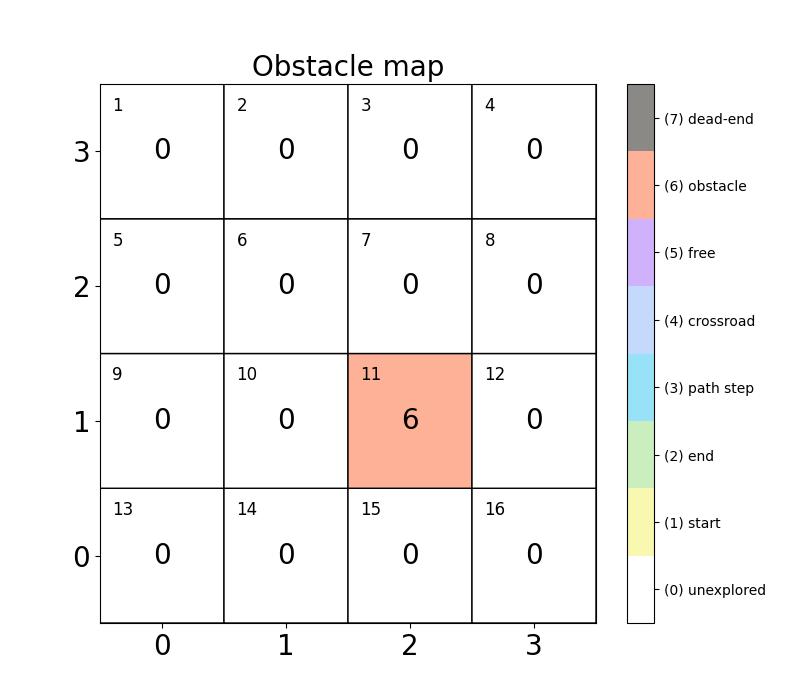}
    \caption{}
    \end{subfigure}%
    \begin{subfigure}{0.33\textwidth}
    \centering
    \includegraphics[width=0.99\textwidth]{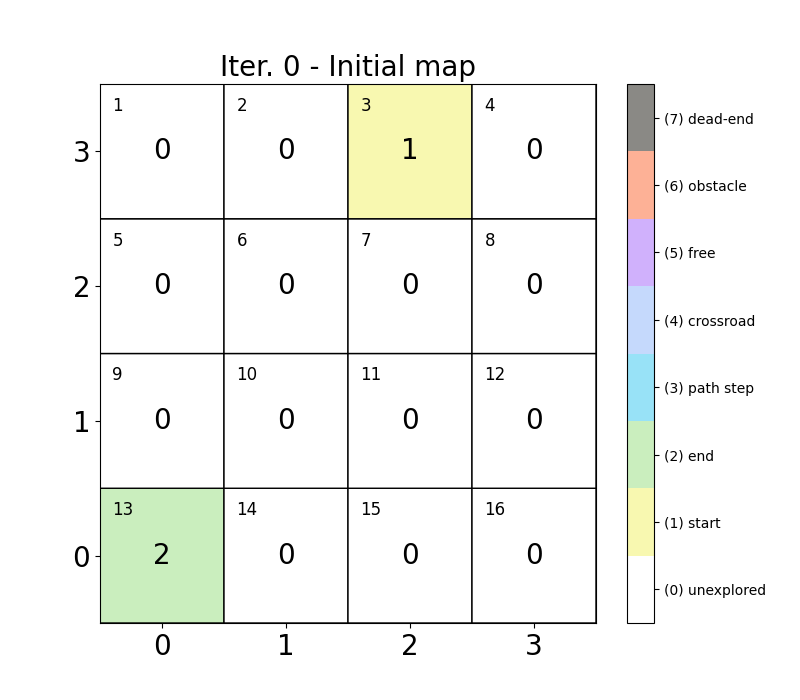}
    \caption{}
    \end{subfigure}%
    \hfill
    \begin{subfigure}{.33\textwidth}
    \centering
    \includegraphics[width=0.99\textwidth]{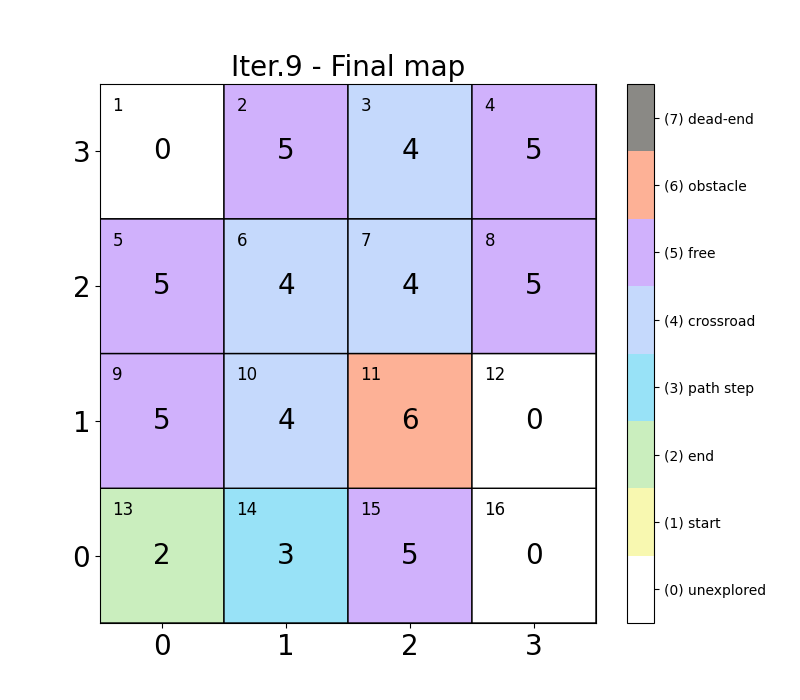}
    \caption{}
    \end{subfigure}
    \caption{State maps of the environment for the 4$\times$4 grid map experiment: (a) showing the location of the obstacle that the system must avoid; (b) at the beginning of the simulation with the start and goal positions; (c) after finishing the simulation when reaching the target.}
    \label{fig:4x4simplemap}
\end{figure}

\begin{figure}[!t]
    \centering
    \begin{subfigure}{.5\textwidth}
    \centering
    \includegraphics[width=\textwidth]{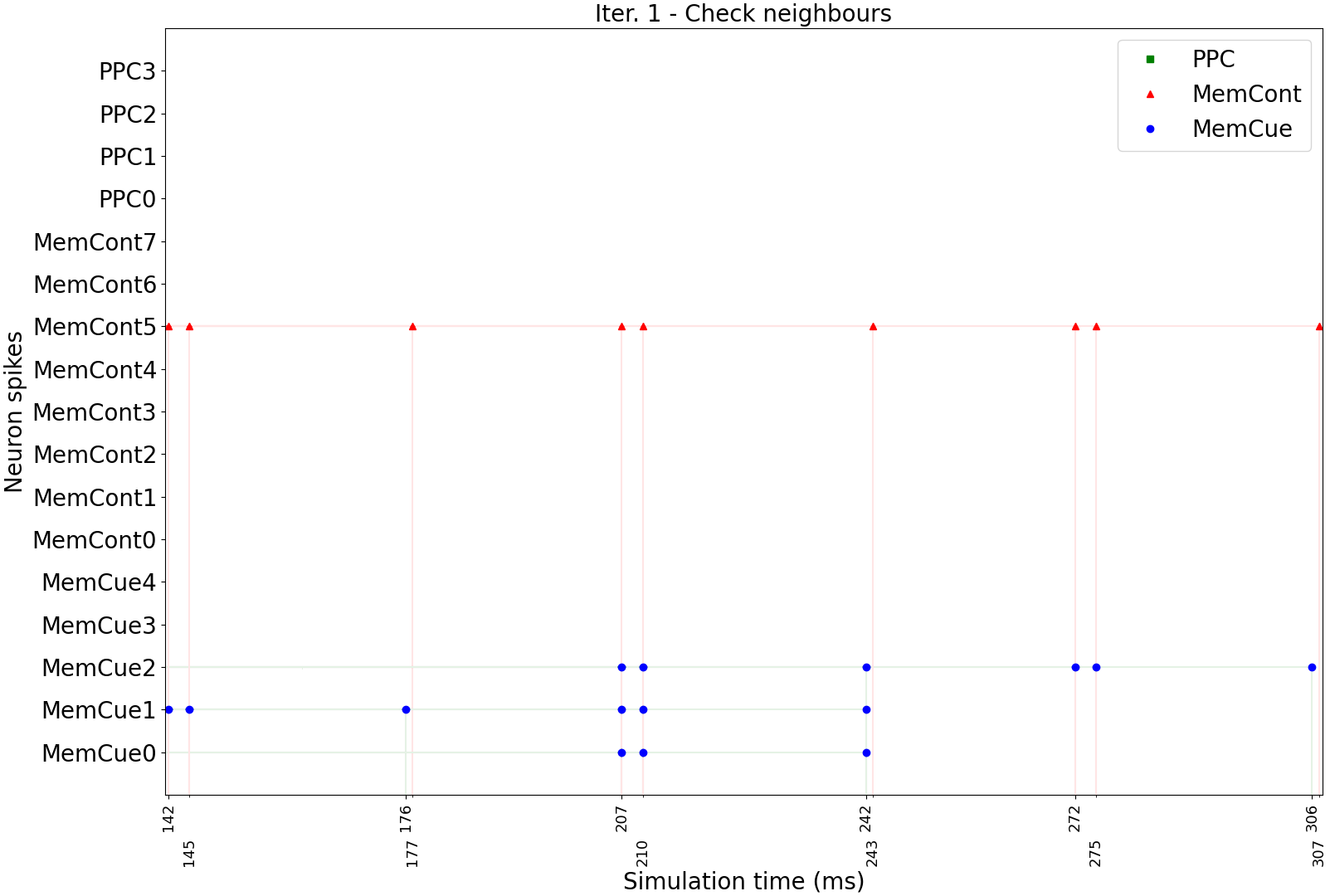}
    \caption{}
    \end{subfigure}
    \par\medskip
    \begin{subfigure}{.5\textwidth}
    \centering
    \includegraphics[width=\textwidth]{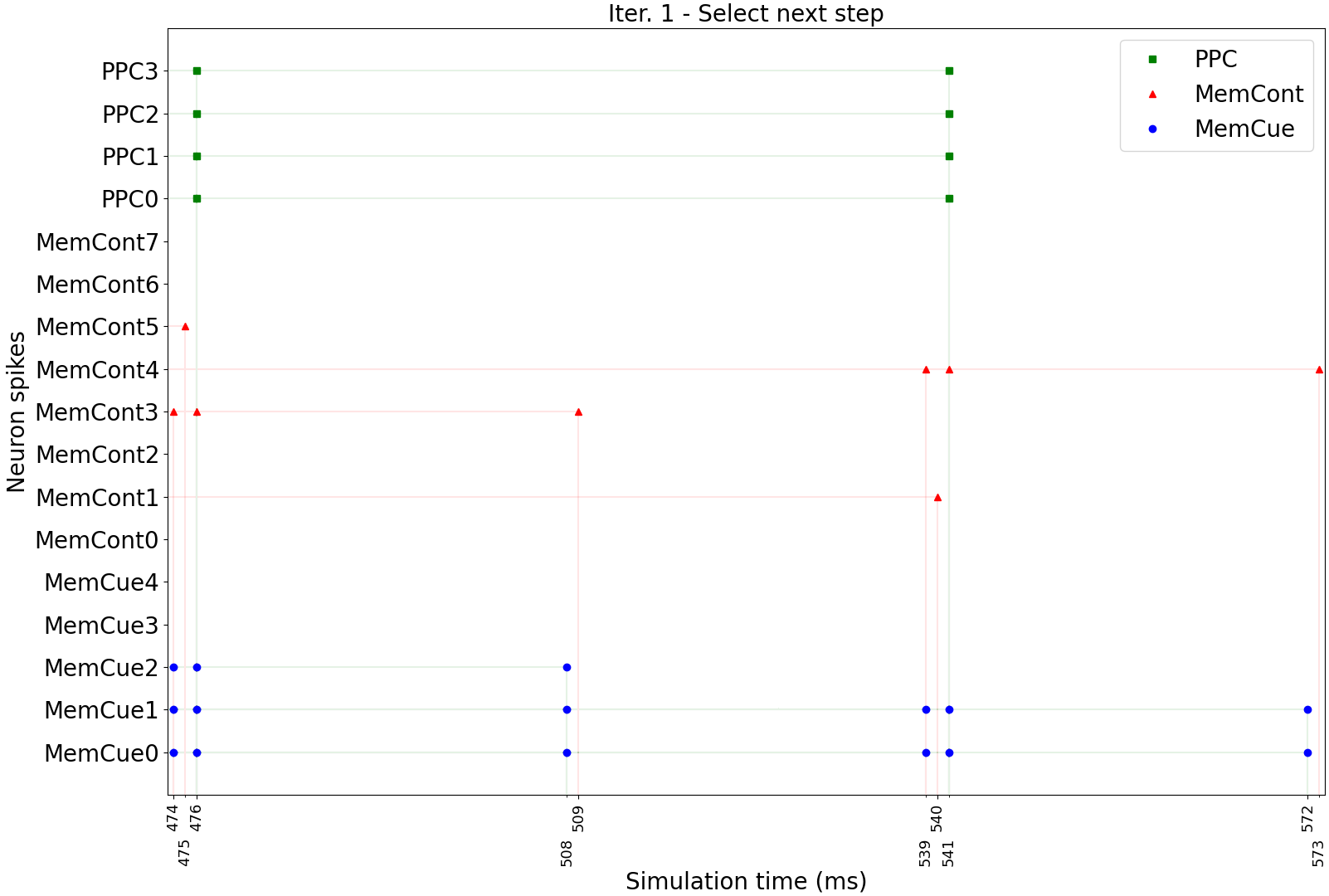}
    \caption{}
    \end{subfigure}
    \par\medskip
    \begin{subfigure}{.5\textwidth}
    \centering
    \includegraphics[width=\textwidth]{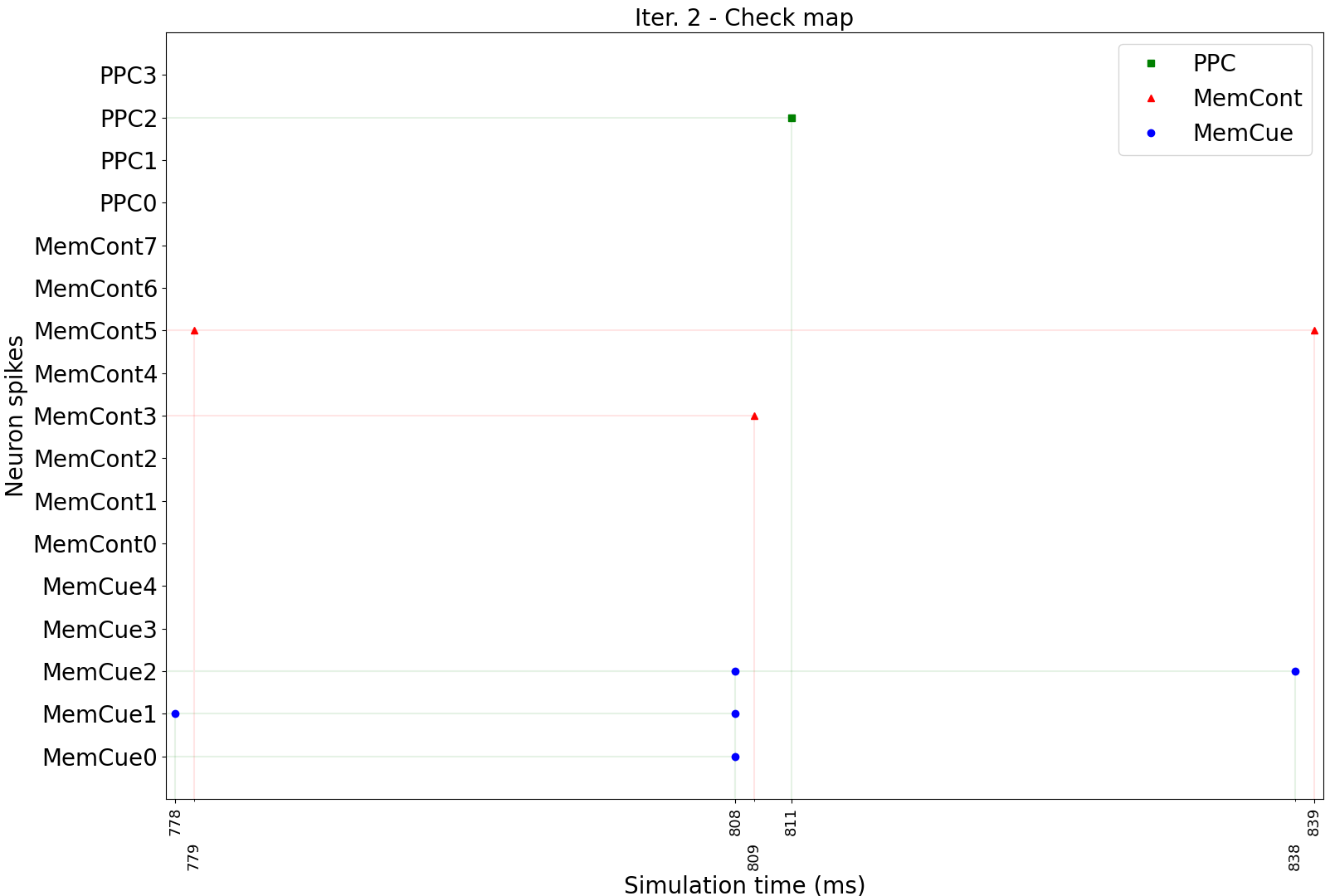}
    \caption{}
    \end{subfigure}
    \caption{Spike activity of the proposed pseudo-mapping and navigation model during the 4$\times$4 grid map experiment: (a) in a local environment mapping phase during the first iteration; (b) when selecting the next position to move to in order to reach the goal during the first iteration; (c) in the decision-making phase during the second iteration. Each point represents the activation of a neuron (Y axis) at a given time instant (X axis). The spikes generated by the PPC are shown in green, those generated by the part of the memory in charge of the memory cue are represented in blue, and those corresponding to the part of the memory in charge of the rest of the memory content are displayed in red.}
    \label{fig:4x4simplespike}
\end{figure}

This is the first experiment with the complete system on a virtual grid environment of 4$\times$4 size, i.e., a total of 16 positions distributed in 4 rows and 4 columns. The environment presents a single obstacle strategically placed on the trajectory taken by the robotic system in the ideal case with no obstacles within the same environment (see Figure~\ref{fig:4x4simplemap}a).

Initially, the system will learn the initial and target positions (Figure~\ref{fig:4x4simplemap}b). The virtual robot will initially be oriented downwards and, according to this map, start at position 3. After that, the iterative execution of the proposed system will start. In each iteration, the system begins with a local environment mapping phase in which measurements of the local environment will be taken to determine whether or not there is an obstacle around the current position.

The real-time spiking activity of the system during this phase in the first iteration is shown in Figure~\ref{fig:4x4simplespike}a. The local environment of the robotic system will be formed by the 3 surrounding positions, i.e., right, front and left. As the robot starts at position 3 looking down, its immediate right position would be 2, the front would be 7 and the left would be 4. For all of them, the state is an obstacle-free position, as there is only one placed at position 11. 

For each position, a state learning operation with reinforcement is performed, i.e., a learning operation followed by a recalling operation of the same memory. The memories will consist of the position they encode as a cue and the state of that position as content.

With these considerations in mind, in Figure~\ref{fig:4x4simplespike}a it is first observed the learning and recalling operations of the state of position 2. Learning is characterized by the activation of neuron 1 of the memory cue population (\textit{MemCue1}), encoding position 2 in binary, and neuron 5 of the memory content population (\textit{MemCont5}), encoding the obstacle-free state, at ms~142 and 145. For the recalling operation, the position is passed as a cue (activation of \textit{MemCue1}) at ms~176 leading to the recall of the state of that position (activation of \textit{MemCont5}) at ms~177.

The same occurs for the remaining neighboring positions at ms~207-243 for position 7 (\textit{MemCue0}, \textit{1} and \textit{2}) and ms~272-307 for position 4 (\textit{MemCue2}), with an obstacle-free state or activation of \textit{MemCont5}. 

For the learning operations, the memory must receive the input memory for 3 consecutive time units. However, only the first and third input stimuli affect the memory; the second input spike is ignored by neurons due to their refractory period. Considering this, for the first operation, for example, if the first spike occurs at ms~142, the next one should be at ms~144. Since the system is working in real time, it may not reach the exact precision of 1 ms between signal transmissions and it may arrive 1~ms later, as in this first learning operation. However, the memory is sufficiently consistent, to ensure that it does not pose a problem and can function correctly even in these cases, as demonstrated by the correct recall of the stored memory during reinforcement learning. 

After this, the decision-making phase takes place, in which the next action to be taken to reach the goal position will be determined or, if there is no neighboring position with the \textit{step in path} state, the most suitable position for being the next in the path will be selected.

In the first iteration, the system is surrounded by free neighboring positions, but none of them has the \textit{step in path} state (see Figure~\ref{fig:4x4simplemap}b). By means of Manhattan distance, positions 7 and 2 are the closest to the target. As there are more than one option, by default, the system will choose the position that reduces the distance on the vertical axis, i.e., position 7. At the same time, since the current position (position 3) has more than one possible path to the target, its state will change to crossroad. Both learning operations with reinforcement are reflected in Figure~\ref{fig:4x4simplespike}b.

Both learning operations with reinforcement are shown at ms 474-509 for position 7 with the \textit{step in path} state (activation of \textit{MemCont3}) and at ms 539-573 for position 3 with crossroad state (activation of \textit{MemCont4}). Both state changes lead to forgetting the previous state of those positions in order to learn the new ones, as denoted by their last recall at ms 475 and 540 for position 7 and 3, respectively.

The activation of the \ac{PPC} outputs at ms 476 and 541 is due to the input activity saturation corresponding to reinforcement learning operations on states of interest, as a constant excitatory input is supplied 2 to 3 times in a row from which only a single arrival is expected at most.

After that, the first iteration would end and the second one would follow. In its decision-making phase, there is a next position in the trajectory to follow, therefore, the PPC generates the motor action necessary to reach it. As shown in Figure~\ref{fig:4x4simplespike}c, the phase starts with the recall of the state of the local environment at ms 778. When \textit{MemCont3} (\textit{step in path} state) related to position 7 arrives at the PPC at ms 809, PPC's output 2 is activated at ms 811. This output corresponds to the motor action of moving down (continuing forward) one position with respect to the current one.

In short, in the first iteration, the local environment is mapped and the next position to be visited is decided, while in the second iteration the motor command to reach that position is generated. In other words, with this pair of iterations, a step is taken on the path to the goal. 

In iterations 3 and 4, the system is at position 7 and detects the obstacle at position 11. This is recorded in the hippocampal memory. The system does not consider that position as the next step in the trajectory and the PPC ignores that position as it has a state that is not of interest, i.e., the position has an obstacle. Therefore, the robotic system successfully avoids it.

From here on, the system will constantly repeat this pair of iterations until the goal is reached at iteration 9. The map of states of the environment stored in the hippocampal memory after reaching the goal can be seen in Figure~\ref{fig:4x4simplemap}c.

With this experiment, the common operation flow of the complete system is described. At the same time, the correct operation of the system is verified for a 4$\times$4 grid environment with a simple obstacle in the trajectory, being able to travel from the source position to the target position avoiding the obstacle and mapping the environment around the trajectory.

\subsection{6$\times$6 grid map simulation with short backtracking experiment}
\label{subsec:6x6simple}

\begin{figure}[!t]
    \centering
    \begin{subfigure}{.33\textwidth}
    \centering
    \includegraphics[width=1\textwidth]{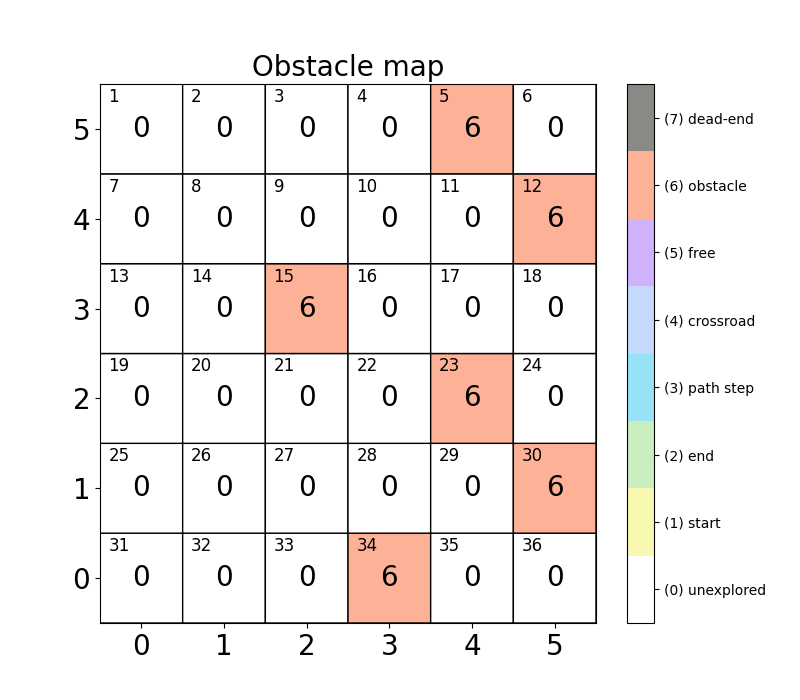}
    \caption{}
    \end{subfigure}%
    \begin{subfigure}{0.33\textwidth}
    \centering
    \includegraphics[width=1\textwidth]{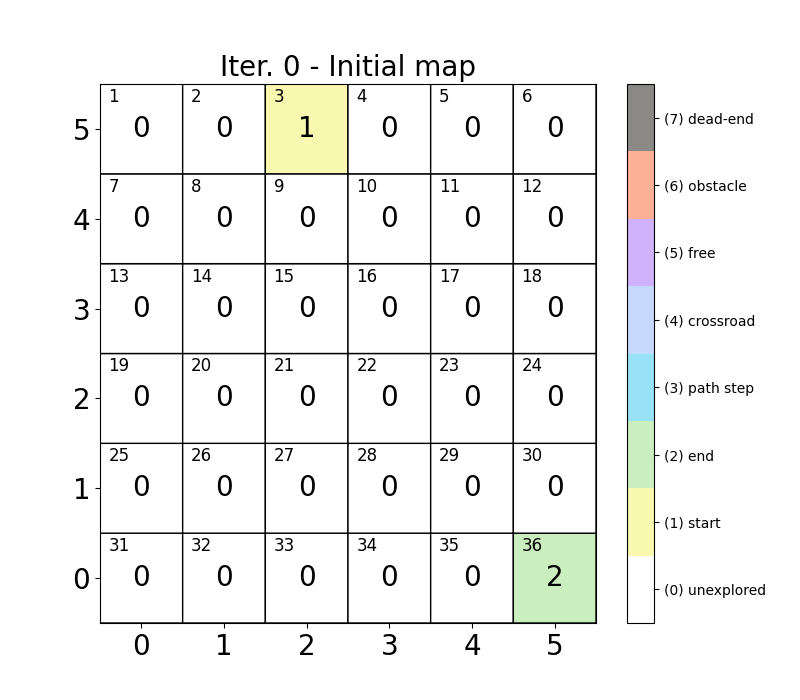}
    \caption{}
    \end{subfigure}%
    \hfill
    \begin{subfigure}{.33\textwidth}
    \centering
    \includegraphics[width=1\textwidth]{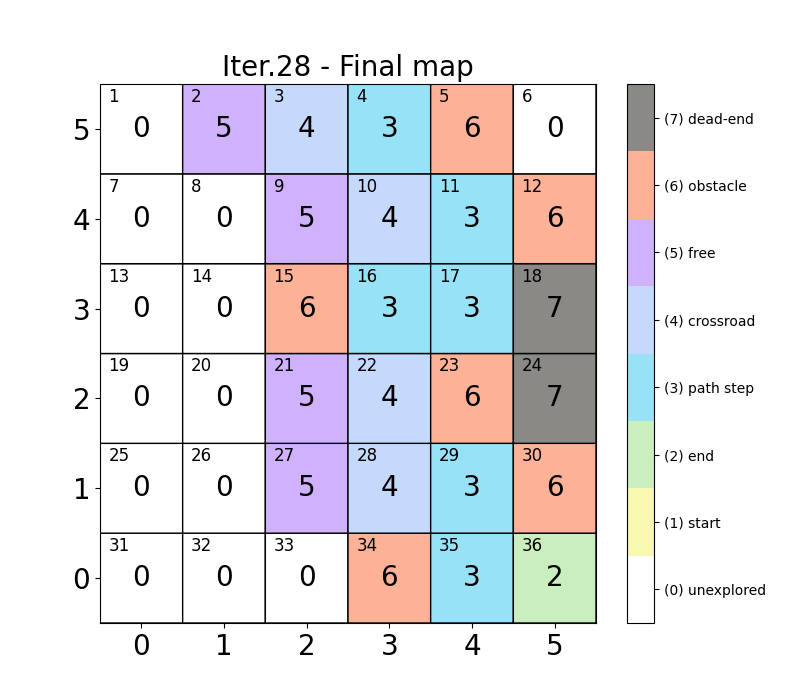}
    \caption{}
    \end{subfigure}
    \caption{State maps of the environment for the 6$\times$6 grid map with short backtracking experiment: (a) showing the location of the obstacle that the system must avoid; (b) at the beginning of the simulation with the start and goal positions; (c) after finishing the simulation when reaching the target.}
    \label{fig:6x6simple}
\end{figure}

In this second experiment regarding the complete system, the aim is to test it in a more complex environment than the previous one in terms of size and number and distribution of obstacles. The system will navigate through a virtual grid environment of 6$\times$6 size, i.e., 36 positions. In this environment there is a total of 6 obstacles strategically placed on the trajectory that the robot would take to reach the goal if no obstacles were present. In this case, the obstacles (see Figure~\ref{fig:6x6simple}a) will lead the robot to a dead-end, where a backtracking mode that was not needed in the previous experiment and that allows the robot to go back to the first crossroad and look for another possible path is used.

The system starts the experiment knowing its initial and target position (Figure~\ref{fig:6x6simple}b). Following its normal operation mode, every two iterations it advances one position in the environment while mapping the state of the neighboring positions. The sequence of positions and actions that the system performs is as follows: position 3 (left), position 4 (right), position 10 (left), position 11 (right), position 17 (left), position 18 (right) and position 24. This way, the robot followed this sort of "ladder movement" to try to reach the goal since, every time it tried to maintain a direction, it encountered an obstacle. 
During this path, the system mapped positions 5, 12, 23 and 30 with the \textit{obstacle} state (\textit{MemCont6} activation), positions 4, 11, 18 and 24, with the \textit{step in path} state (\textit{MemCont3} activation), positions 3, 10 and 17 with the \textit{crossroad} state (\textit{MemCont4} activation), since they allowed more than one possible path to the goal, and the remaining positions surrounding the followed path that had no obstacles were assigned the \textit{obstacle free} state (\textit{MemCont5} activation).
When position 24 is reached, the system does not find any position in the local environment that is free of obstacles, i.e., it reached a dead-end. At this point, the system starts its backtracking mode. In this mode, the system turns the orientation of the robot's head by 180°. In other words, the robot turns around, and begins to follow the path it previously took in search for a position with crossroad state. A position with this state means that there is at least one other free position in its local environment that has not been explored yet.

In this way, the system turns around and moves forward through position 18 until it reaches position 17. This backtracking consists in following the positions of the local environment with \textit{step in path} state that are already mapped in the hippocampal map representation. Therefore, to step through each position, only a single iteration is required: in the same iteration the state of the local positions is recalled and the PPC generates the necessary motor action to proceed. At the same time, all the positions traversed until reaching the one with the crossroad state are marked with the \textit{dead-end} state (activation of \textit{MemCont7}) to indicate that these positions only lead to dead-end paths.

At position 17, the system returns again to the normal behavior mode, continuing its search for the next position towards the target, giving preference to those positions that have not been visited yet. When passing through position 17 again, as it has only one possible position left to continue that was not visited, its state is changed from \textit{crossroad} to \textit{step in path}. Following the normal mode of operation by pairs of iterations, the system reaches the target position at iteration 28. The state map of the environment stored in the hippocampal memory after the simulation can be seen in Figure~\ref{fig:6x6simple}c. Position 14 is not marked as a crossroad because the system reaches the target position and terminates the execution, and therefore it does not update the position.

This experiment demonstrates the correct behavior of the system in a larger environment and with a more complex obstacle placement, forcing the robot to temporarily switch to backtracking mode, where it stays for a couple of iterations until reaching a crossroad position.

\subsection{6x6 grid map simulation with long backtracking experiment}
\label{subsec:6x6complex}

\begin{figure}[!t]
    \centering
    \begin{subfigure}{.33\textwidth}
    \centering
    \includegraphics[width=1\textwidth]{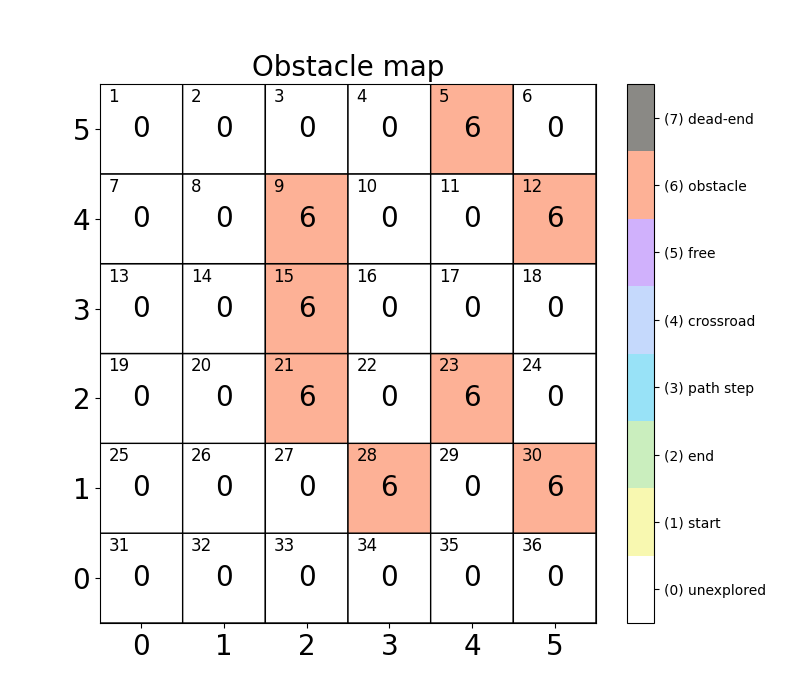}
    \caption{}
    \end{subfigure}%
    \begin{subfigure}{0.33\textwidth}
    \centering
    \includegraphics[width=1\textwidth]{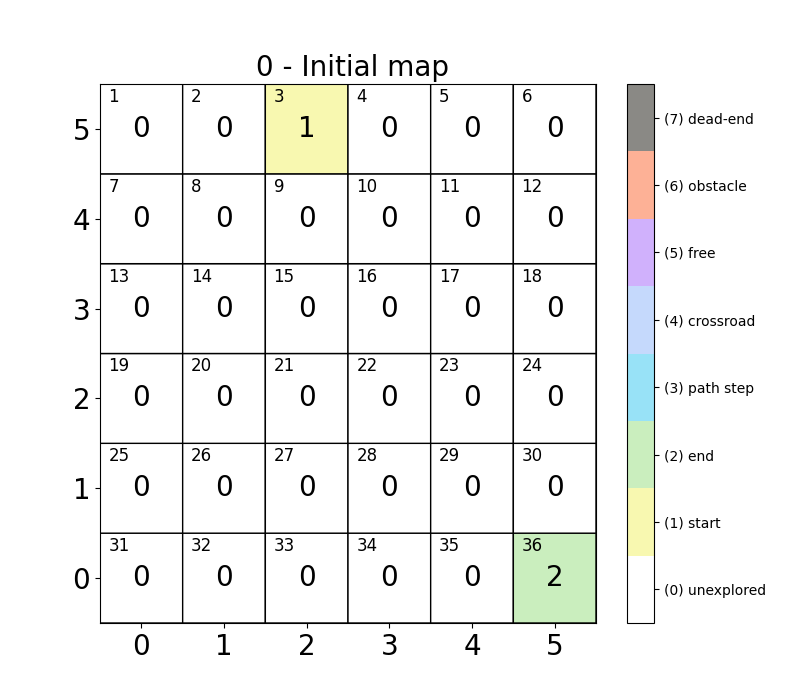}
    \caption{}
    \end{subfigure}%
    \hfill
    \begin{subfigure}{.33\textwidth}
    \centering
    \includegraphics[width=1\textwidth]{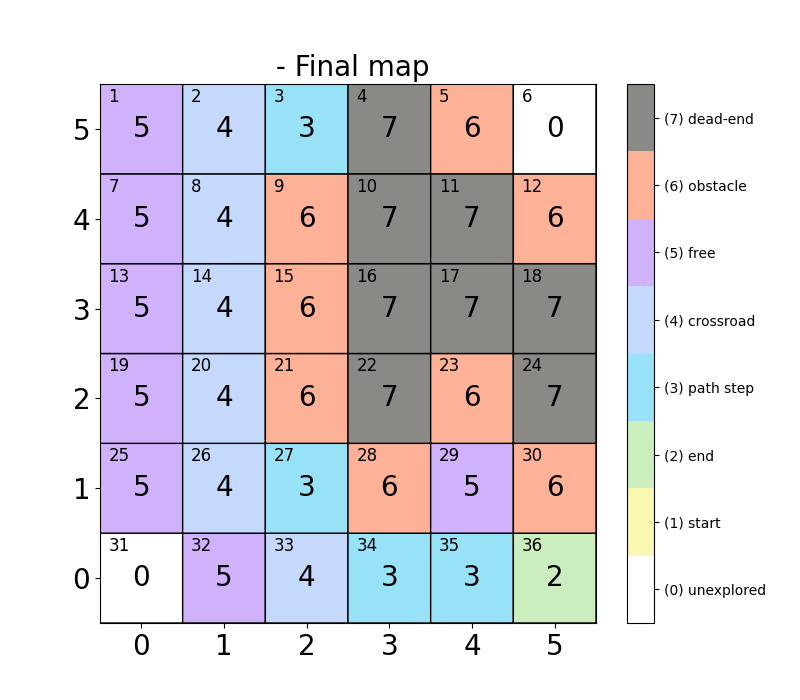}
    \caption{}
    \end{subfigure}
    \caption{State maps of the environment for the complex 6$\times$6 grid map experiment: (a) showing the location of the obstacle that the system must avoid; (b) at the beginning of the simulation with the start and goal positions; (c) after finishing the simulation when reaching the target.}
    \label{fig:6x6complex}
\end{figure}


In this experiment, the system navigates through a virtual environment of 6$\times$6 size with a large number of obstacles than in the previous experiment. These obstacles (see Figure~\ref{fig:6x6complex}a) prevent the system from reaching the goal through the set of ideal shortest paths (right side of the environment). The system goes around these obstacles, traversing almost the entire environment, in order to reach the goal position.

At the beginning, the system follows the same sequence of actions as in the previous experiment. The difference is  present when reaching position 17 after backtracking, where it continues its trajectory through position 16 and arrives at position 22, where the system again encounters another dead-end. Unlike in the previous experiment, the backtracking path it has to perform involves more iterations and backtracking movements. Specifically, it completely explores the whole right part of the environment until it returns to the initial position. In this process, the \textit{dead-end} state is assigned to all the positions in that region. 

After that, it takes the path to the left side of the environment despite moving away from the goal, in order to explore other possible paths that will allow the robot to reach it. On this path, it avoids the central obstacle wall until reaching the goal position without further problems. After reaching this position, the hippocampal memory presents an almost complete state map of the environment containing the shortest path between the initial position and the goal (see Figure~\ref{fig:6x6complex}c).

This demonstrates the correct operation of the complete system in both normal and backtracking modes, as well as its ability to map the environment, even in complex and relatively large environments.

\subsection{Real-time demonstration using a robotic platform}

\begin{figure}[!t]
    \centering
    \includegraphics[width=0.55\textwidth]{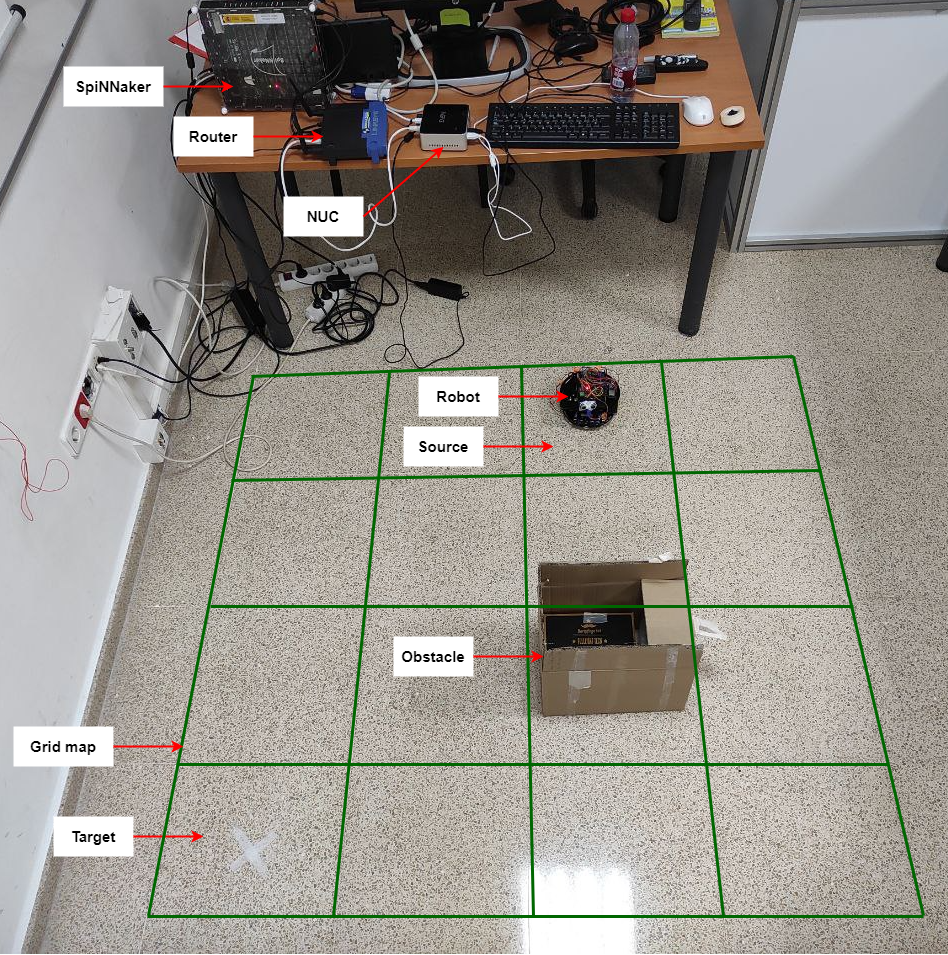}
    \caption{Environment used for the real-time demonstration with the robotic platform. It contains the grid map with the source and target positions of the simulation, the obstacle and the necessary hardware systems. These hardware systems are the robotic platform, the SpiNNaker board, the computer (NUC) that keeps the whole system running and the router needed to interconnect the NUC with SpiNNaker.}
    \label{fig:setup}
\end{figure}

The previous experiments demonstrated the operation of the system in virtual environments, simulating the inputs and outputs. This experiment was performed in order to prove the operation of the whole system in a real scenario. A real-time demonstration of the system operation in a real physical environment was developed using the robotic platform described in Section~\ref{sec:materials_and_methods}.

For this experiment, the same grid as the one used in the "4$\times$4 grid map simulation experiment" was used. Therefore, the same initial and goal positions in a 4$\times$4 grid environment and a single obstacle placed in the path of the robot towards the goal were used (see Figure~\ref{fig:setup}). In order to prepare this environment, the physical space was divided into cells of 30$\times$30 centimeters, resulting in an environment of 120$\times$120 centimeters. 

The SpiNNNaker hardware platform is used to run the \ac{SNN} formed by the hippocampal memory and the \ac{PPC}. The NUC acts as the core system to launch the network simulation on SpiNNaker, manage the iterative execution of the system and the communication between the network and the robotic platform. This communication consists in passing the information related to the presence or absence of an obstacle in the local environment obtained from the robot to the \ac{SNN} implemented in SpiNNaker, and the response from SpiNNaker with the motor action generated by the PPC to the robot in order to execute it.

After the experiment was finished, the robotic platform was able to reach the target position while avoiding the obstacle. The spiking system generated exactly the same spiking activity and map as in the 4$\times$4 grid virtual environment that used the same configuration. In short, the operation of the system was demonstrated not only in ideal virtual environments, but also in physical and real case environments.

A recording of this experiment is available, with annotations of what happens at each moment. This video can be found in the repository indicated at the end of Section~\ref{sec:conclusions}.
\section{Discussion}
\label{sec:discussion}

The set of experiments performed demonstrated the decision-making capacity of the PPC model to generate the motor actions from the state inputs of the different positions in the local environment. On the other hand, they proved the operation of the complete system in different environments, both virtual with simulated inputs and physical with a robotic platform in a real case scenario. In addition, these experiments were used to test the system progressively in environments whose complexity ranged from lower to higher, varying the size of the environment and the number and disposition of the obstacles in it.

Each experiment on the complete system was used to explore different situations where its ability to adapt to the environment was demonstrated, both in navigation to reach the goal by detecting and avoiding obstacles and in the mapping of the explored environment. In the first experiments, since the environments used were simple, the system barely explored and mapped half of it. On the other hand, in the last experiment consisting of a virtual environment, the system mapped almost the entire environment due to its level of complexity and the need to search for alternative paths to reach the goal position.

The hippocampal memory model shows a certain degree of forgetting for each operation performed (memory leak) \cite{casanueva2022bio}. This forgetting factor could be adjusted, by modifying the STDP parameters, to add a temporality factor to the validity of the information. In this way, local and more recent information would be retained for longer, while information from more distant positions in both time and space would be forgotten, i.e., they would lose validity. This mechanism would allow the system to reinforce its performance in dynamic environments. When navigating through already visited positions that were forgotten, the system will map them again.


From a bio-inspired point of view, the biological basis of the PPC (Section~\ref{subsec:biological_background}) describes in broad outline the functions in which it participates, as well as some of the internal structures that form it. When designing and developing the proposed PPC model, we tried to make it as close to these bases as possible.

In the biological model, the PPC actively participates in the navigation process by transforming a sequence of allocentric information from the hippocampus into a sequence of egocentric information, such as action sequences. On the other hand, the proposed model takes the information of the state of the local environment (i.e., the surrounding positions) as input to determine the egocentric motor action (turn right, turn left, up and down) to be performed in each iteration. In addition, the local environment state information comes from an allocentric representation of the environment via a map stored and managed by the hippocampal memory model.

The PPC model generates motor actions at each iteration (or pair of iterations), where each iteration represents the location of the system at a different position within a discrete environment. In other words, the PPC generates motor actions at discrete positions as in the biological model. 

At the structural level, in the biological model of the PPC, Match-selective neurons can be found. Each set of these neurons is responsible for generating a specific motor action based on the information coming from sensory flows and the flow of brain regions with more abstract cognitive processes. Similarly, the proposed model presents a population of Match-selective neurons, each of which are responsible for generating a specific motor action. This population determines the action to be performed based on the information from the states of interest of the local positions where the system is located. States that derive directly from external sensory information or from the processing of these.

Although the state of the positions directly come from the map constructed in the hippocampal memory, the information with which these states are assigned comes both from the sensory flow (absence or presence of obstacles, target, etc.) and from regions with more complex cognitive processes (step-in-path, dead-end, etc.). In addition, the proposed model presents a population of inhibitory interneurons that are responsible for resetting the state of the Match-selective neurons (which is also present in the biological model) after each iteration.

The biological model of the hippocampus indicates that it acts as a short-term memory and is also capable of maintaining an allocentric representation of the environment. This representation is achieved by place cells, which are neurons that are activated when the mammal is in a certain position in the discretized environment. In the proposed system, the hippocampus acts as a memory capable of storing memories that form an allocentric representation of the environment. This is achieved by considering that memory cues encode the activity of the place cells (the positions in the environment), and that the content represents the possible states of those positions.

As discussed in Section~\ref{subsubsec:mapping}, the states of positions come from neurons and brain regions whose activity is received at the hippocampus. The states related to the presence of obstacles, absence of obstacles, initial position and goal come from the Border and Goal cells, while the states of step in path, crossroad and dead-end would come from those brain regions involved in more abstract processes, such as semantic memory.

The complete system presents a normal behavior when searching for new positions that bring it closer to the goal and a backtracking behavior in which it searches for new possible paths for reaching the goal position. This has also been observed in biology. Authors in \cite{whitlock2008navigating} and \cite{oess2017computational} indicate that the functioning of the PPC and the hippocampus is sensitive to the general behavior of the individual. It is therefore plausible that, during navigation, depending on the individual's situation, one behavior or another is adopted in these brain regions.

However, there are some aspects of the model that are not similar to its biological counterpart. There is a set of features yet to be implemented by \acp{SNN} that are beyond the scope and purpose of this work. The mapping of the local environment and the local decision-making or path planning are performed by the proposed spiking network. While the management signals of the system iterations, the global planning or deciding which is the closest position to the goal of those found with the local environment mapping, and the communication with the robot are external to the spiking network.

The bio-inspired nature of the proposed system comes from its inspiration in the brain regions of the hippocampus and the PPC, not in the application of \ac{ANN}-to-SNN conversion methods, as in \cite{fischl2017neuromorphic}. The proposed system has not only been simulated in software and on ideal environments as in \cite{tang2018gridbot} and \cite{nakagawa2022neural}, but also on more complex environments and, in addition, on a robotic platform in physical environments.

Regarding the environment, the proposed system initially knows nothing about the environment except the initial and goal position. Throughout the simulation, the system maps the environment through a continuous learning process. Therefore, it does not work with completely-known and previously-mapped environments where no learning is involved, as in \cite{oess2017computational} and \cite{sakurai2021path} or those that focus only on path-planning and, therefore, assume that they can access the status of the entire map, such as \cite{koziol2014neuromorphic}, \cite{hwu2017adaptive}, \cite{friedrich2016goal} and \cite{koul2019waypoint}.

The hippocampus model is responsible for cumulatively storing the result of the local mapping, resulting in a pseudomapping of the environment, while in \cite{nichols2010case} the mapping result is not stored, but immediately used for local planning. This map presents different states beyond the presence of obstacles or not and the goal position, as in \cite{nakagawa2022neural}, \cite{kreiser2018pose} and \cite{tang2018gridbot}. In addition, the system reacts to the environment by avoiding obstacles and reaching the goal, not merely circling around the environment avoiding obstacles or even ignoring them as in \cite{nakagawa2022neural} and \cite{tang2018gridbot}.

The proposed system does not present a global planning algorithm achieved with SNNs like those proposed in \cite{koziol2014neuromorphic}, \cite{hwu2017adaptive}, \cite{friedrich2016goal}, \cite{koul2019waypoint} and \cite{sakurai2021path}. However, these techniques necessarily require knowledge of the entire environment and additional information that is sometimes difficult to obtain. In \cite{kreiser2018pose}, \cite{kreiser2020error} and \cite{tang2019spiking} the authors propose SNN models to approximately determine the position of the robot in each iteration of the system to achieve a SLAM, while the proposed system requires external signals to control the position in each iteration. Moreover, \cite{nakagawa2022neural}, \cite{kreiser2018pose} and \cite{tang2018gridbot} use a more bio-inspired position state input system than that of the proposed system.
\section{Conclusions}
\label{sec:conclusions}

In this work, a fully functional spike-based robotic navigation and environment pseudomapping system implemented with SNNs on the SpiNNaker hardware platform was proposed. The proposed system is able to navigate through a discrete and initially unknown grid environment in order to reach a goal position from an initial position while avoiding possible obstacles that may be present in the trajectory. Furthermore, the system is able to map the state of the explored environment during its trajectory to the goal.

The architecture of the proposed system is fully parameterized, thus allowing its use in environments of different sizes or even in environments of equal size but with higher or lower spatial resolution in the discretization of the environment. Specifically, it demonstrated good performance for environments of up to 36 positions distributed in a 6x6 grid with many obstacles, but it would work just as well for other maps of different sizes, as well as different number and distribution of obstacles.

In this work, a fully functional bio-inspired PPC neural network model implemented with SNNs was also proposed. This model is able to perform local decision-making or planning based on the state of the positions in the local environment. This information regarding the state of the environment is provided by a hippocampal memory model proposed in a previous work.

This memory model is able to maintain a representation of the state of the different positions that make up the grid-like environment through which the system navigates. To do this, it learns and recalls memories that consist of a cue (fragment from which the rest of the content of the memory is recalled), and the content of the memory itself. Each memory encodes the activation of a place cell (i.e., a position on the map of the environment) through the cue, while the rest of the content of the memory indicates the state of that position. The information with which the possible states are established comes from the Border cells, Goal cells and other regions with higher abstraction functionalities.

Furthermore, the proposed complete system was not only compared with its biological counterpart, but also with other systems that can be found in the literature. A bio-inspired spiking system capable of mapping the environment that the robot navigates with a wide range of possible states for each position was proposed. Thanks to this range of states, the system was able to navigate complex environments by detecting and avoiding obstacles until reaching the goal position. In addition, a PPC model capable of performing local decision-making or planning based on information about the state of the local environment was presented. 

The proposed functionalities of both the complete system and the bio-inspired PPC model were demonstrated through a set of experiments. These experiments were successfully completed by demonstrating, on the one hand, the decision-making capability of the PPC model, and, on the other hand, the ability of the complete system to navigate and map a set of progressively larger and more complex environments. Furthermore, its operation was not only simulated, but also tested in a physical environment with a robotic platform in a real-time demonstration.  

As a future work, some spiking features of the system could be implemented. One of such features would be the adaptation of the wavefront technique for the hippocampal memory architecture as a spiking global planning algorithm. This allows the system to react to dynamic environments by having a global planning that determines the trajectory to the goal and a local planning that rectifies depending on the obstacles it detects. Another interesting feature would be the incorporation of a population of head cells with which to determine in a spiking manner the position in which the system is at any given moment and, therefore, which are the neighboring positions.

In addition, the spiking map generated in hippocampal memory after reaching the goal has many applications in the field of neuromorphic robotics. It could be shared with other spiking robotic systems that, without the need for learning and mapping the environment, could navigate from the source position to the goal or calculate the most optimal route to the goal. Several robots in parallel could explore different regions of an environment and, after reaching their respective goals, assemble a joint spiking representation of their explored regions. Ultimately, the generation of the spiking state map of the environment and even the map itself serve as the basis for spiking robotic swarms.

The source code regarding the implementation of the complete spike-based robotic navigation and environment pseudomapping system with \acp{SNN} on SpiNNaker is available on GitHub (\url{https://github.com/dancasmor/Bio-inspired-spike-based-Hippocampus-and-Posterior-Parietal-Cortex-robotic-system-for-pseudo-mapping}).

\medskip

\medskip
\textbf{Acknowledgements} \par 
This research was partially supported by the Spanish grant MINDROB (PID2019-105556GB-C33/\\AEI/10.13039/501100011033). D. C.-M. was supported by a "Formaci\'{o}n de Profesorado Universitario" Scholarship from the Spanish Ministry of Education, Culture and Sport.

\medskip
\bibliographystyle{IEEEtran}
\bibliography{bibliography}

\end{document}